\documentclass{article} 
\usepackage{iclr2027_conference,times}

\usepackage{hyperref}
\usepackage{url}

\usepackage{graphicx}
\usepackage{wrapfig}

\usepackage{amsmath}
\usepackage{amssymb}
\usepackage{amsthm}
\usepackage{algorithm}
\usepackage{algorithmic}
\usepackage{multirow}
\usepackage{booktabs}
\usepackage[table]{xcolor}
\usepackage{xcolor}
\usepackage{makecell}
\usepackage{enumitem}

\usepackage{booktabs}
\usepackage{graphicx}
\usepackage[table]{xcolor}
\usepackage{caption}

\newtheorem{proposition}{Proposition}

\definecolor{green}{RGB}{0,150,0}
\definecolor{red}{RGB}{200,0,0}

\title{Orthogonal Yet Coupled: Decoupling Geometric Components for Model Merging}

\iclrfinalcopy

\author{
Zijing Wang$^{1}$,
Yongkang Liu$^{1}$\thanks{Corresponding authors.} ,
Mingyang Wang$^{2,3}$,
Ercong Nie$^{4}$,
Mengjie Zhao$^{1}$\\
\textbf{Yunpu Ma$^{2,3}$,}
\textbf{Kang Liu$^{1}$,}
\textbf{Zihan Wang$^{1}$,}
\textbf{Shi Feng$^{1}$,}
\textbf{Daling Wang$^{1}$}\footnotemark[1] ,
\textbf{Hinrich Schütze$^{2,3}$} \\
\AND
\normalfont\mdseries
$^{1}$Northeastern University, China \\
$^{2}$CIS, LMU Munich, Germany \\
$^{3}$Munich Center for Machine Learning (MCML), Germany \\
$^{4}$Shanghai Jiao Tong University, China \\
\texttt{wzj1718@gmail.com}
}

\usepackage{xspace}
\newcommand{\methodname}{DiGA\xspace}

\begin{document}

\maketitle

\begin{abstract} 
Merging pretrained models has emerged as an effective approach for consolidating diverse capabilities into a single unified model. However, prevailing merging methods typically treat each task vector as an indivisible merging unit, overlooking the heterogeneous geometric changes encoded within it. This treatment can induce \emph{cross-component coupling}: when merging decisions are derived from statistics of the complete task vector, the geometric characteristics of one component may influence how another is selected, weighted, or combined, potentially degrading the quality of the merged model. To address this issue, we propose \textbf{DiGA}, a \textbf{Di}sentangled \textbf{G}eometry-\textbf{A}ware model merging framework. Using the pretrained weights as a shared geometric reference, DiGA orthogonally decomposes each task vector into components corresponding to distinct geometric attributes. Rather than merging the task vectors as a whole, DiGA aggregates corresponding components independently within their respective subspaces and subsequently recombines them into a unified update. This component-wise formulation preserves the geometric identity of each component and prevents the characteristics of one component from interfering with the aggregation of another. Furthermore, DiGA can be incorporated into a broad range of existing model merging methods. Extensive experiments across diverse models, tasks, and merging methods demonstrate that DiGA improves merged-model performance and reduces capability degradation. Our repository is on~\url{https://github.com/wzj1718/DiGA}.
\end{abstract}

\section{Introduction}
\label{sec:introduction}

Artificial General Intelligence (AGI) represents the ultimate goal of AI research, requiring models capable of performing a diverse range of tasks across various domains~\citep{legg2007universal,bommasani2021opportunities,wei2026designing}. The computational cost of pre-training large models for AGI from scratch has become prohibitively high~\citep{kaplan2020scaling,hoffmann2022training}. A promising alternative is to build upon existing pretrained models and develop specialized experts through fine-tuning, which can then be consolidated to acquire diverse capabilities. Model merging offers a scalable approach to consolidating domain-specific fine-tuned models into a single unified model without additional training~\citep{mitchellwortsman2022model,wang2025scaling,yang2026model}.

Most existing model merging methods operate on \emph{task vectors}~\citep{ilharco2023taskarithmetic,huang2024emr,chen2025bring}, defined as the parameter differences between task-specific fine-tuned models and a shared pretrained model.
Early approaches, such as Task Arithmetic~\citep{ilharco2023taskarithmetic}, merge task-specific knowledge through linear combinations of task vectors, with scaling coefficients controlling their relative contributions.
Subsequent methods improve upon direct superposition by sparsifying task vectors to retain task-relevant parameters~\citep{yu2024dare}, resolving conflicting update signs across models~\citep{yadav2023ties}, 
or exploiting low-rank structures to selectively retain dominant update directions before aggregation~\citep{lee2025star,stoica2025model}. 
Recent studies have further examined the geometric relationships among task vectors and leveraged orthogonality on the orthogonal-group manifold to alleviate interference during merging~\citep{yang2026orthomerge}.

Despite these advances, existing methods generally treat each task vector as a single, indivisible merging unit. This treatment implicitly assumes that all changes encoded in a task update can be handled by the same aggregation rule. However, a task vector may simultaneously encode multiple types of geometric changes relative to the pretrained parameters~\citep{qiu2023oft,ma2024parameter}. In particular, a task update can contain changes in \textbf{orientation} and \textbf{structure}. The former reorients the pretrained parameter structure, whereas the latter modifies its internal geometry or extends beyond it. Yet existing merging methods do not explicitly distinguish these effects, instead operating on the complete task vector as a single geometric object.

We argue that such treatment introduces a previously overlooked form of interference, which we term \textbf{cross-component coupling}. 
Many merging methods derive masks, weights, signs, normalization factors, or other aggregation decisions from statistics of the complete task vector. 
Consequently, the magnitude, direction, or distribution of one geometric component can alter the aggregation decisions applied to another, even when the two components occupy orthogonal subspaces. 
Geometrically distinct changes that should ideally be handled independently become coupled during merging. This source of interference is distinct from the commonly studied conflicts among task vectors from different experts: it arises \emph{within} each task vector, before or during cross-model aggregation.

To address this issue, we propose \textbf{DiGA}, a \textbf{Di}sentangled \textbf{G}eometry-\textbf{A}ware model merging framework. DiGA uses the pretrained model as a shared geometric reference and orthogonally decomposes each task vector into orientation and structural components. 
Instead of merging the original task vectors directly, DiGA aggregates the two component families independently within their respective subspaces and then recombines them into a unified update. 
By preventing the statistics of one component family from affecting the aggregation of another, DiGA explicitly eliminates cross-component coupling while retaining the original semantics of the underlying merging operator.

\textbf{Contributions.}
(i) We identify \textbf{cross-component coupling}, a previously overlooked interference in model merging that arises when geometrically heterogeneous components within a task vector are jointly processed by the same aggregation decisions.
(ii) We propose \textbf{DiGA}, which orthogonally decomposes each task vector into orientation and structural components, independently merges corresponding components within their respective subspaces, and subsequently recombines them into a unified update.
(iii) We extensively evaluate DiGA across diverse models, tasks, and merging methods, demonstrating consistent improvements in merged-model performance and broad compatibility with existing merging frameworks.

\section{Background}
\label{sec:motivation}

\subsection{Task-Vector Merging}
\label{sec:task_vector_merging}

Let $\theta_0$ denote a pretrained model and
$\{\theta_i\}_{i=1}^{N}$ denote task-specific experts obtained by
independently fine-tuning $\theta_0$. For a matrix parameter with base
weight $W_0$ and corresponding expert weight $W_i$, we define the task
update as
\begin{equation}
\Delta_i = W_i - W_0.
\label{eq:task_update}
\end{equation}
Task-vector merging forms the merged weight by aggregating these updates:
\begin{equation}
W^\star = W_0 + \mathcal{F}(\{\Delta_i\}_{i=1}^{N}),
\label{eq:generic_merging}
\end{equation}
where $\mathcal{F}$ denotes the merging operator. Although existing
methods instantiate $\mathcal{F}$ differently, they generally treat each
$\Delta_i$ as a single and indivisible aggregation unit.

\subsection{Limitations of Whole-Vector Aggregation}
\label{sec:whole_vector_limitation}

Treating a task vector as a single aggregation unit implicitly assumes
that all changes encoded within it can be processed under the same
merging decisions. However, fine-tuning can induce heterogeneous geometric
changes relative to the shared pretrained parameters. Some changes may
primarily alter the orientation of the pretrained parameter structure,
whereas others may modify its internal geometry or introduce variation
beyond the pretrained structure.

Whole-vector aggregation does not explicitly distinguish these effects.
This becomes particularly problematic for adaptive merging methods, whose
masks, weights, sign decisions, normalization factors, or expert-selection
rules are derived from statistics of the complete task update. Under such
operators, the geometric characteristics associated with one type of change
can affect how another type is aggregated. We refer to this phenomenon as
\emph{cross-component coupling}.

This observation motivates two questions: how can the heterogeneous
geometric effects within a task update be explicitly disentangled, and how
can they be merged without allowing the characteristics of one component
to influence the treatment of another? We address these questions with
\methodname.

\section{Methodology}
\label{sec:method}

\methodname is a disentangled geometry-aware model merging framework that
uses the pretrained model as a shared geometric reference. It first
constructs an orthogonal decomposition of each task update into components
with distinct geometric roles, then aggregates corresponding components
independently within their respective subspaces, and finally recombines
the resulting updates into a unified model.

\subsection{Geometric Disentanglement of Task Updates}
\label{sec:geometric_disentanglement}

Since all experts originate from the same pretrained model, the base weight
provides a natural shared reference for characterizing their geometric
changes. We use this reference to distinguish an \emph{orientation
component}, which captures structure-preserving changes in the orientation
of the pretrained parameter frame, from a \emph{structural component},
which captures changes that alter the internal geometry of the frame or
extend beyond the pretrained parameter subspace.

For $W_0\in\mathbb{R}^{m\times n}$ with $m\geq n$ and full
column rank, its reduced QR factorization \citep{bjorck1967solving} is
\begin{equation}
W_0=QH, \qquad Q^\top Q=I,
\label{eq:base_qr_revised}
\end{equation}
where $Q\in\mathbb{R}^{m\times n}$ has orthonormal columns spanning the
column space of $W_0$, and $H\in\mathbb{R}^{n\times n}$ is an upper
triangular matrix. The columns of $Q$ provide a common coordinate frame
for all expert updates. We take $Q$ itself as the geometric anchor. Its
in-frame tangent directions define a reference subspace for decomposing
task updates.


The first-order directions associated with within-frame reorientation of
the orthonormal frame $Q$ have the form $QA$, where the coordinate matrix
$A$ is skew-symmetric, i.e., $A^\top=-A$
(proof in Appendix~\ref{app:reorientation_component}). These directions
form the \emph{orientation subspace}:
\begin{equation}
\mathcal{T}_Q=\{QA:A^\top=-A\}.
\label{eq:reorientation_subspace_revised}
\end{equation}
This subspace captures structure-preserving changes in the orientation of
the pretrained parameter frame.
To obtain an exact orthogonal decomposition of task updates, we pair
$\mathcal T_Q$ with its orthogonal complement:
\begin{equation}
\mathcal{R}_Q=\mathcal{T}_Q^\perp
=\{X\in\mathbb{R}^{m\times n}:Q^\top X\ \text{is symmetric}\}.
\label{eq:reshaping_subspace_revised}
\end{equation}
Here, $X$ denotes a change in the original parameter space, and $Q^\top X$
gives its coordinates within the reference frame. We refer to
$\mathcal R_Q$ as the \emph{structural subspace}, as it captures changes
that modify the internal geometry of the pretrained frame or introduce
variation outside its span. The characterization of $\mathcal{R}_Q$ in
Eq.~\eqref{eq:reshaping_subspace_revised} is proved in
Appendix~\ref{app:structure_altering_subspace}.

The two subspaces impose different in-frame coordinate conditions:
skew-symmetric for $\mathcal{T}_Q$ and symmetric for $\mathcal{R}_Q$.
To obtain the corresponding components of a task update $\Delta_i$, we
therefore express its in-frame coordinates as
$G_i=Q^\top\Delta_i$ and separate them into skew-symmetric and symmetric
parts. This gives the unique orthogonal decomposition
$G_i=\operatorname{skew}(G_i)+\operatorname{sym}(G_i)$, where
\begin{equation}
\operatorname{skew}(G_i)=\frac{G_i-G_i^\top}{2},\qquad
\operatorname{sym}(G_i)=\frac{G_i+G_i^\top}{2}.
\label{eq:skew_sym_definitions_revised}
\end{equation}

\begin{proposition}[Base-Induced Orthogonal Decomposition]
\label{prop:base_decomposition_revised}
For a fixed base frame $Q$, the matrix space admits the orthogonal direct sum
\begin{equation}
\mathbb{R}^{m\times n}=\mathcal{T}_Q\oplus^\perp\mathcal{R}_Q.
\label{eq:direct_sum_revised}
\end{equation}
Here, $\oplus^\perp$ denotes an orthogonal direct sum.
Consequently, every task update has the unique decomposition
\begin{equation}
\Delta_i=T_i+R_i,\qquad
T_i=P_{\mathcal{T}_Q}(\Delta_i),\quad
R_i=P_{\mathcal{R}_Q}(\Delta_i),
\label{eq:additive_decomposition_revised}
\end{equation}
where $P_{\mathcal{T}_Q}$ and $P_{\mathcal{R}_Q}$ denote orthogonal
projections under the Frobenius inner product. Their closed forms are
\begin{align}
T_i
&=Q\operatorname{skew}(G_i)
=Q\operatorname{skew}(Q^\top\Delta_i),
\label{eq:reorientation_component_revised}\\
R_i
&=Q\operatorname{sym}(G_i)+(I-QQ^\top)\Delta_i.
\label{eq:reshaping_component_revised}
\end{align}
For any pair of experts $i$ and $j$, this decomposition further gives
\begin{equation}
\langle T_i,R_j\rangle_F=0.
\label{eq:decomposition_consequences_revised}
\end{equation}
\end{proposition}

The existence, uniqueness, and cross-component orthogonality of the
decomposition are proved in
Appendix~\ref{app:unified_decomposition_proof}. For $m<n$, we use the
analogous row-space construction in
Appendix~\ref{app:wide_decomposition}.
We refer to $T_i$ and $R_i$ as the \emph{orientation component} and
\emph{structural component}, respectively. 
In $R_i$, $Q\operatorname{sym}(G_i)$ captures within-frame structural deformation, while $(I-QQ^\top)\Delta_i$ captures variation beyond the pretrained parameter span.
Because $Q$ is shared across experts, corresponding components lie in the same geometric subspaces. 
Both components are substantial: Appendix
Table~\ref{tab:oft_lora_geometry} reports average ratios
$\|T_i\|_F/\|R_i\|_F$ of $0.809$ for OFT and $0.520$ for LoRA.
Their orthogonality, however, does not guarantee independent processing by a merging operator.

\begin{figure*}[t]
\centering
\includegraphics[width=\linewidth]
{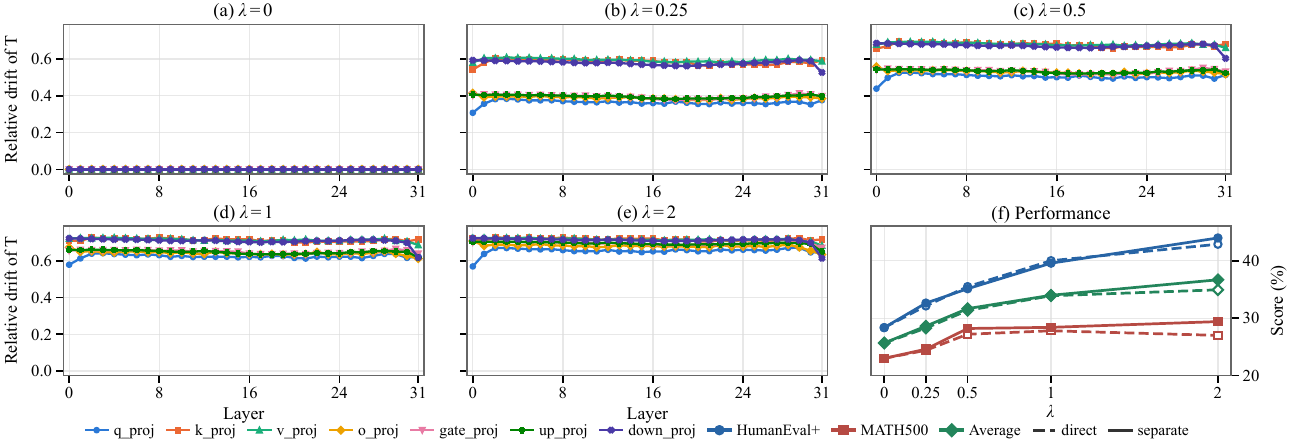}
\caption{
\textbf{Cross-component coupling and performance in TIES-Merging.}
Using five OFT experts based on Llama-3.1-8B, we vary the structural
contribution as $\Delta_i(\lambda)=T_i+\lambda R_i$ while keeping $T_i$
fixed.
(a--e) For each $\lambda$, both routes use identical magnitude masks
computed from the complete task updates.
$T_{\mathrm{mix}}(\lambda)$ applies the sign-election and expert-selection
decisions computed from $\{\Delta_i(\lambda)\}_{i=1}^{N}$ to the fixed
orientation components, whereas $T_{\mathrm{sep}}(\lambda)$ computes these
decisions from $\{T_i\}_{i=1}^{N}$ under the same masks.
We report the layer-wise relative drift
$\|T_{\mathrm{mix}}(\lambda)-T_{\mathrm{sep}}(\lambda)\|_F/
\|T_{\mathrm{sep}}(\lambda)\|_F$.
The increasing drift shows that the structural contribution alters how
the fixed orientation components are aggregated.
(f) Corresponding performance of direct and separate TIES merging across
$\lambda$ on HumanEval+ and MATH500, together with their arithmetic mean.
Separate processing achieves a higher average at every tested nonzero
$\lambda$.
}
\label{fig:coupled}
\end{figure*}

\subsection{Cross-Component Coupling}
\label{sec:component_coupling}

The decomposition above formalizes the limitation of whole-vector
aggregation described in Section~\ref{sec:whole_vector_limitation}.
For a fixed linear merger, decomposing each task update as
$\Delta_i=T_i+R_i$ does not change the merged result, since linear
aggregation distributes over the two components. This equivalence
generally breaks for adaptive merging rules whose decisions depend on
update statistics. By Proposition~\ref{prop:base_decomposition_revised},
\begin{equation}
\|\Delta_i\|_F^2 = \|T_i\|_F^2+\|R_i\|_F^2, \qquad
\langle\Delta_i,\Delta_j\rangle_F
=
\langle T_i,T_j\rangle_F+\langle R_i,R_j\rangle_F .
\end{equation}
Thus, despite cross-component orthogonality, statistics of the complete
task updates still mix information from both component families. When
such statistics govern adaptive merging decisions, one component can
influence how the other is selected, weighted, or combined, establishing
the \emph{cross-component coupling} described in
Section~\ref{sec:whole_vector_limitation}.

TIES-Merging~\citep{yadav2023ties} makes this dependence particularly
visible. Its elected sign and retained expert set are
determined from the complete task updates; consequently, changing $R_i$
can change how a fixed $T_i$ is aggregated.
Figure~\ref{fig:coupled} quantifies this effect through a controlled
structural-strength sweep. 
As the contribution of $R_i$ increases, the
aggregation of the fixed $T_i$ under whole-vector decisions increasingly drifts from its independently aggregated counterpart. The corresponding
performance results further show that this coupling can affect the quality of the merged model. These observations motivate preserving
component identity throughout the merging process.

\subsection{Component-Wise Aggregation}
\label{sec:component_aggregation}

To eliminate cross-component coupling, \methodname aggregates corresponding
components independently within their respective geometric subspaces.
Specifically, the same branch operator $\mathcal{G}$ is independently
applied to the orientation and structural component families:
\begin{equation}
\Delta_T
=\mathcal{G}\left(\{T_i\}_{i=1}^{N}\right),
\qquad
\Delta_R
=\mathcal{G}\left(\{R_i\}_{i=1}^{N}\right).
\label{eq:basin_component_aggregation_restructured}
\end{equation}
The resulting component-wise updates are then recombined with the
pretrained weight:
\begin{equation}
W^\star=W_0+\Delta_T+\Delta_R.
\label{eq:basin_reconstruction_restructured}
\end{equation}

Because $\mathcal{G}$ computes aggregation statistics independently within
each component family, one family cannot affect the aggregation decisions
of the other. Thus, \methodname preserves component identity and avoids
cross-component coupling from whole-vector adaptive aggregation.

This disentangle--aggregate--recombine formulation is not tied to a
specific merging rule: compatible operators can be applied independently
to the two component families, as examined in
Section~\ref{sec:sr_generality}. Beyond this general
framework, 
\methodname instantiates $\mathcal{G}$ to account for directional redundancy, source strength, and directional consensus within each subspace.
For notational convenience, let $\{D_i\}_{i=1}^{N}$ denote either
$\{T_i\}_{i=1}^{N}$ or $\{R_i\}_{i=1}^{N}$. \methodname constructs a
nonredundant aggregate direction while preserving source strengths, then
calibrates its magnitude using directional consensus.

\paragraph{Redundancy-aware direction construction.}
Expert components within the same geometric subspace can overlap
directionally, causing direct aggregation to repeatedly reinforce shared
directions and distort the merged update. Prior work shows that such
repeated aggregation can inflate dominant singular values and degrade
performance~\citep{li2026when}; Appendix Figure~\ref{fig:cos} also shows
nontrivial cosine similarity within each component family.
To reduce this redundancy while preserving source strength, we decompose
each component into its norm $n_i$ and unit orientation $u_i$:
\begin{equation}
n_i=\|D_i\|_F,
\qquad
u_i=\frac{D_i}{n_i}.
\label{eq:source_norm_direction_restructured}
\end{equation}
We retain $n_i$ and apply Gram--Schmidt orthogonalization only to the orientations.
Let $\mathcal A_i=\{j<i:\|r_j\|_F>\epsilon\}$ denote the directions retained
before processing source $i$. 
We compute
\begin{equation}
r_i=u_i-\sum_{j\in\mathcal A_i}
\langle u_i,e_j\rangle_F e_j,
\qquad
e_i=\frac{r_i}{\|r_i\|_F}\ \text{if }\|r_i\|_F>\epsilon.
\label{eq:component_gram_schmidt_restructured}
\end{equation}
Here, $r_i$ is the part of $u_i$ unexplained by previously retained
directions. After processing all sources, let
$\mathcal A=\{i:\|r_i\|_F>\epsilon\}$ denote the retained set. We then
combine these directions using the original source norms:
\begin{equation}
v=\sum_{i\in\mathcal A}n_i e_i,
\qquad
\hat v=\frac{v}{\|v\|_F}.
\label{eq:merged_direction_restructured}
\end{equation}
The resulting $\hat v$ is a nonredundant aggregate direction that preserves
the relative influence of the original expert components. Numerical details
and ordering sensitivity are reported in
Appendix~\ref{app:diga}.

\paragraph{Consensus-aware magnitude calibration.}

The unit direction $\hat v$ specifies the orientation of the component-wise
update but does not determine how far the merged model should move along
this direction. We use the mean source norm
$\bar n=\frac{1}{N}\sum_{i=1}^{N}n_i$
as a reference adaptation scale.

Source strength alone, however, does not indicate whether the experts
consistently support a common direction. We therefore measure directional
consensus using the average pairwise cosine similarity of the original
normalized components:
\begin{equation}
\bar c
=\frac{1}{N(N-1)}
\sum_{i\neq j}\langle u_i,u_j\rangle_F.
\label{eq:branch_consensus_restructured}
\end{equation}
We compute $\bar c$ before orthogonalization because the original overlap,
although redundant when constructing the aggregate direction, provides the
relevant signal of agreement among experts. We use the following
parameter-free affine calibration:
\begin{equation}
\gamma=(1+\bar c)\bar n.
\label{eq:branch_magnitude_restructured}
\end{equation}
This is the simplest rule that retains the mean source scale when average
similarity is zero, increases it under positive agreement, and attenuates
it under negative agreement. Combining this magnitude with the aggregated
direction gives
\begin{equation}
\mathcal{G}\left(\{D_i\}_{i=1}^{N}\right)
=\gamma\hat v
=(1+\bar c)\bar n\,\hat v.
\label{eq:branch_aggregation_restructured}
\end{equation}
Applied independently to the orientation and structural component families,
this operator yields $\Delta_T$ and $\Delta_R$ in
Eq.~\eqref{eq:basin_component_aggregation_restructured}. Their subsequent
orthogonal recombination through
Eq.~\eqref{eq:basin_reconstruction_restructured}
produces the final \methodname merged model.

\begin{table*}[t]
\centering
\scriptsize
\caption{
Performance of merging five Llama-3.1-8B experts adapted with LoRA or OFT.
}
\label{tab:lora_oft_results}
\resizebox{\textwidth}{!}{
\begin{tabular}{llccccc|c|cc|c}
\toprule
& Model & MATH500 & HumanEval+ & ScienceQA & CommonsenseQA & Social-IQA
& Task Avg. & AGIEval & ARC-Avg & Transfer Avg. \\
\midrule

& Llama-3.1-8B
& 18.40 & 22.44 & 71.27 & 70.60 & 48.26
& 46.19 & 35.38 & 40.27 & 37.82 \\
\midrule

\multirow{13}{*}{LoRA}

& Individual Experts
& 27.20 & 41.28 & 90.87 & 83.70 & 58.14
& 60.24 & -- & -- & -- \\

& TA
& 26.60 & 38.48 & 86.47 & 80.43 & 54.40
& 57.28 & 38.27 & 43.00 & 40.64 \\

& OrthoMerge-C
& 25.60 & 37.74 & 86.69 & 80.10 & 54.35
& 56.90 & \textbf{38.43} & 43.04 & \textbf{40.74} \\

& OrthoMerge-G
& 24.60 & 38.66 & 86.42 & 79.69 & 54.66
& 56.81 & 37.47 & 41.33 & 39.40 \\

& TIES
& \textbf{27.60} & 41.59 & 83.45 & 80.59 & 56.19
& 57.88 & 38.05 & 42.92 & 40.49 \\

& OrthoMerge-C
& 27.20 & \textbf{43.05} & 83.86 & 81.16 & 58.24
& 58.70 & 37.72 & 42.50 & 40.11 \\

& OrthoMerge-G
& 24.60 & 38.54 & 84.08 & 79.85 & 56.40
& 56.69 & 36.90 & 41.12 & 39.01 \\

& TSVM
& 19.40 & 33.60 & 87.86 & \textbf{83.13} & \textbf{58.55}
& 56.51 & 33.62 & 41.22 & 37.42 \\

& OrthoMerge-C
& 19.80 & 31.59 & 88.04 & \textbf{83.13} & 58.44
& 56.20 & 34.26 & 41.54 & 37.90 \\

& OrthoMerge-G
& 21.80 & 41.04 & 88.53 & 82.56 & 57.68
& 58.32 & 36.39 & 41.30 & 38.85 \\

& \cellcolor{gray!15}\textbf{\methodname}
& \cellcolor{gray!15}26.40
& \cellcolor{gray!15}40.49
& \cellcolor{gray!15}\textbf{88.89}
& \cellcolor{gray!15}82.31
& \cellcolor{gray!15}57.83
& \cellcolor{gray!15}\textbf{59.18}
& \cellcolor{gray!15}36.16
& \cellcolor{gray!15}\textbf{43.23}
& \cellcolor{gray!15}39.70 \\

\midrule

\multirow{7}{*}{OFT}

& Individual Experts
& 27.20 & 38.78 & 91.28 & 82.56 & 56.76
& 59.32 & -- & -- & -- \\

& TA
& 25.20 & 32.93 & 83.45 & 76.74 & 51.89
& 54.04 & 38.47 & 41.98 & 40.22 \\

& TIES
& 27.80 & 40.00 & 81.88 & 77.07 & 52.35
& 55.82 & 37.83 & 41.99 & 39.91 \\

& DARE
& 18.80 & 31.28 & 83.72 & \textbf{80.92} & \textbf{58.03}
& 54.55 & 34.70 & 38.37 & 36.53 \\

& TSVM
& 23.40 & 35.91 & 86.38 & 80.18 & 55.02
& 56.18 & 36.82 & 42.34 & 39.58 \\

& OrthoMerge
& 24.80 & 38.41 & 87.72 & 80.51 & 55.17
& 57.32 & 38.78 & 42.75 & 40.76 \\

& \cellcolor{gray!15}\textbf{\methodname}
& \cellcolor{gray!15}\textbf{29.40}
& \cellcolor{gray!15}\textbf{40.12}
& \cellcolor{gray!15}\textbf{88.26}
& \cellcolor{gray!15}79.93
& \cellcolor{gray!15}55.27
& \cellcolor{gray!15}\textbf{58.60}
& \cellcolor{gray!15}\textbf{38.97}
& \cellcolor{gray!15}\textbf{42.86}
& \cellcolor{gray!15}\textbf{40.92} \\

\bottomrule
\end{tabular}
}
\end{table*}

\section{Experiments and Results}
\label{sec:experiments}

\subsection{Experimental Setup}
\label{sec:experimental_setup}
We evaluate \methodname against representative data-free merging methods,
including Task Arithmetic (TA)~\citep{ilharco2023taskarithmetic}, TIES-Merging~\citep{yadav2023ties}, DARE~\citep{yu2024dare}, TSVM~\citep{gargiulo2025tsv}, and OrthoMerge-C/G~\citep{yang2026orthomerge}, across both
language-only and vision-language settings. Our experiments cover
parameter-efficient experts adapted with LoRA and OFT, fully fine-tuned
language experts, and vision-language experts specialized in spatial
reasoning, OCR, and medical multimodal reasoning. We apply \methodname to
the two-dimensional attention and MLP projection matrices in each
Transformer block, while parameters outside this decomposition are merged
using TA. Evaluation covers the corresponding target capabilities together
with transfer benchmarks where applicable.

We defer the complete experimental specifications
to the appendix. Appendix~\ref{app:experimental_setup} provides the expert
models, task configurations, and detailed merging setup;
Appendix~\ref{app:bench} describes the evaluation benchmarks;
Appendix~\ref{app:merge_method} summarizes the compared merging methods; and
Appendix~\ref{app:implementation_details} specifies the evaluation harnesses,
decoding and scoring protocols, expert orderings, and configuration.

\subsection{Main Results}
\label{sec:main_results}

\textbf{Language-only expert merging.}
Table~\ref{tab:lora_oft_results} reports the results of merging five
Llama-3.1-8B experts adapted with LoRA and OFT. \methodname obtains the
highest target-task average in both settings, reaching $59.18\%$ and
$58.60\%$, respectively. These scores exceed the strongest corresponding
baselines by $0.48$ and $1.28$ points and approach the individual-expert
averages, with gaps of only $1.06$ and $0.72$ points. In the OFT setting, it obtains the strongest
MATH500, HumanEval+, and ScienceQA results while remaining competitive on
CommonsenseQA and Social-IQA. These gains persist across LoRA and OFT,
suggesting that geometric disentanglement is not specific to a single adaptation scheme.
\methodname also maintains transfer performance beyond the expert training
domains. Its transfer averages exceed the base-model average of $37.82\%$
in both settings, and it achieves the highest ARC-Avg under both LoRA and
OFT. 
This indicates that improved target-capability integration does not
systematically trade off against the evaluated transfer performance.

{
\setlength{\textfloatsep}{3pt}

\begin{table*}[t]
\centering
\scriptsize
\setlength{\tabcolsep}{2.2pt}
\renewcommand{\arraystretch}{1.12}

\begin{minipage}[t]{0.41\textwidth}
\centering

\captionof{table}{
Merging five fully fine-tuned language experts from MergeBench.
}
\label{tab:general_results}

\resizebox{\linewidth}{!}{
\begin{tabular}{lccccc|c}
\toprule
Method & Multilingual & Coding & Math & Instruction & Safety & Avg \\
\midrule

Llama-3.2-3B
& 34.10 & 27.09 & 28.35 & 6.84 & 32.58 & 25.79 \\

Individual Experts
& 35.00 & 44.29 & 69.83 & 39.93 & 83.38 & 54.49 \\

\midrule

TA
& \underline{35.20} & 37.55 & 40.56 & 10.91 & 41.12 & 33.07 \\

OrthoMerge-C
& \underline{35.20} & 37.32 & 39.73 & 9.61 & 41.00 & 32.57 \\

OrthoMerge-G
& 34.90 & 38.32 & 43.75 & 13.31 & 43.19 & 34.69 \\

TIES
& \underline{35.20} & 40.42 & \underline{56.18}
& \textbf{26.43} & 44.70 & 40.59 \\

OrthoMerge-C
& 35.10 & 40.45 & 55.27 & 22.55 & 43.98 & 39.47 \\

OrthoMerge-G
& 34.90 & 39.39 & 51.48 & 22.74 & 42.53 & 38.21 \\

TSVM
& \textbf{35.30} & \textbf{41.59} & 56.03 & 20.15
& \underline{50.49} & 40.71 \\

OrthoMerge-C
& \textbf{35.30} & 41.09 & \textbf{56.71} & 20.70
& \textbf{50.57} & \underline{40.87} \\

OrthoMerge-G
& 34.90 & 40.71 & 52.77 & 20.33 & 46.12 & 38.97 \\

\rowcolor{gray!15}
\textbf{\methodname}
& \underline{35.20} & \underline{41.34} & 55.22
& \underline{25.32} & 48.37 & \textbf{41.09} \\

\bottomrule
\end{tabular}
}
\end{minipage}
\hfill
\begin{minipage}[t]{0.55\textwidth}
\centering

\captionof{table}{
Merging three fully fine-tuned Qwen2.5-VL-7B-Instruct experts.
}
\label{tab:mm_results}

\resizebox{\linewidth}{!}{
\begin{tabular}{lccccc|c}
\toprule
Method & OCRBench & MMSI-Bench & EmbSpatial & MMMU$_{Med}$ & PathVQA & Avg \\
\midrule

Qwen2.5VL-7B-Instruct
& 84.40 & 27.90 & 69.45 & 53.33 & 66.54 & 60.32 \\

Individual Experts
& 84.60 & 34.20 & 70.16 & 54.67 & 66.72 & 62.07 \\

\midrule

TA
& \textbf{84.50} & 29.60 & 70.80 & 50.00 & 66.12 & 60.20 \\

OrthoMerge-C
& \textbf{84.50} & 29.40 & 70.96 & 50.67 & 66.27 & 60.36 \\

OrthoMerge-G
& 83.70 & 30.70 & 70.99 & 50.67 & 66.48 & 60.51 \\

TIES
& 82.30 & 32.10 & 71.70 & \underline{52.67} & 65.68 & 60.89 \\

OrthoMerge-C
& 83.10 & \textbf{33.10} & 71.90 & \underline{52.67}
& 66.95 & 61.54 \\

OrthoMerge-G
& 84.10 & 32.00 & \textbf{72.42} & \underline{52.67}
& 66.89 & 61.62 \\

TSVM
& 83.30 & 31.00 & 72.09 & 51.33 & \underline{67.55} & 61.05 \\

OrthoMerge-C
& 83.30 & 31.50 & 71.90 & 52.00 & 67.22 & 61.18 \\

OrthoMerge-G
& 83.30 & 32.10 & \underline{72.31} & \textbf{53.33}
& 67.28 & \underline{61.66} \\

\rowcolor{gray!15}
\textbf{\methodname}
& \underline{84.20} & \underline{32.60} & 72.14
& \underline{52.67} & \textbf{67.61} & \textbf{61.84} \\

\bottomrule
\end{tabular}
}
\end{minipage}
\end{table*}

}

\begin{table}[t]
\centering
\caption{
Ablation study of \methodname in the five-expert OFT setting.
Values in parentheses denote changes relative to the complete method.
}
\label{tab:ablation}
\resizebox{\linewidth}{!}{
\begin{tabular}{lcccccc|c}
\toprule
Setting & MATH500 & HumanEval+ & ScienceQA & CommonsenseQA & Social-IQA
& Task Avg. & Transfer Avg. \\
\midrule

\rowcolor{gray!15}
\methodname
& 29.40 & 40.12 & 88.26 & 79.93 & 55.27 & 58.60 & 40.92 \\

\textit{T only}
& 25.00 \textcolor{green}{(-4.40)}
& 28.29 \textcolor{green}{(-11.83)}
& 80.26 \textcolor{green}{(-8.00)}
& 75.35 \textcolor{green}{(-4.58)}
& 50.61 \textcolor{green}{(-4.66)}
& 51.90 \textcolor{green}{(-6.70)}
& 39.86 \textcolor{green}{(-1.06)} \\

\textit{R only}
& 24.60 \textcolor{green}{(-4.80)}
& 35.24 \textcolor{green}{(-4.88)}
& 86.33 \textcolor{green}{(-1.93)}
& 78.62 \textcolor{green}{(-1.31)}
& 53.02 \textcolor{green}{(-2.25)}
& 55.56 \textcolor{green}{(-3.04)}
& 40.50 \textcolor{green}{(-0.42)} \\

\textit{Whole-vector aggregation}
& 28.00 \textcolor{green}{(-1.40)}
& 40.00 \textcolor{green}{(-0.12)}
& 88.17 \textcolor{green}{(-0.09)}
& 79.93 \textcolor{black}{(0.00)}
& 55.12 \textcolor{green}{(-0.15)}
& 58.24 \textcolor{green}{(-0.36)}
& 40.86 \textcolor{green}{(-0.06)} \\

\textit{w/o redundancy removal}
& 28.40 \textcolor{green}{(-1.00)}
& 38.60 \textcolor{green}{(-1.52)}
& 87.72 \textcolor{green}{(-0.54)}
& 80.75 \textcolor{red}{(+0.82)}
& 55.58 \textcolor{red}{(+0.31)}
& 58.21 \textcolor{green}{(-0.39)}
& 40.82 \textcolor{green}{(-0.10)} \\

\textit{w/o consensus calibration}
& 28.20 \textcolor{green}{(-1.20)}
& 40.06 \textcolor{green}{(-0.06)}
& 88.31 \textcolor{red}{(+0.05)}
& 79.93 \textcolor{black}{(0.00)}
& 54.86 \textcolor{green}{(-0.41)}
& 58.27 \textcolor{green}{(-0.33)}
& 40.76 \textcolor{green}{(-0.16)} \\

\textit{w/o source magnitude}
& 24.60 \textcolor{green}{(-4.80)}
& 34.39 \textcolor{green}{(-5.73)}
& 78.15 \textcolor{green}{(-10.11)}
& 81.90 \textcolor{red}{(+1.97)}
& 56.86 \textcolor{red}{(+1.59)}
& 55.18 \textcolor{green}{(-3.42)}
& 39.97 \textcolor{green}{(-0.95)} \\

\textit{Symmetric orthogonalization}
& 29.40 \textcolor{black}{(0.00)}
& 39.76 \textcolor{green}{(-0.36)}
& 86.51 \textcolor{green}{(-1.75)}
& 80.34 \textcolor{red}{(+0.41)}
& 55.12 \textcolor{green}{(-0.15)}
& 58.23 \textcolor{green}{(-0.37)}
& 40.71 \textcolor{green}{(-0.21)} \\

\bottomrule
\end{tabular}
}
\end{table}

We next evaluate \methodname on the fully fine-tuned language experts from
MergeBench. As shown in Table~\ref{tab:general_results}, \methodname obtains
an overall average of $41.09\%$, compared with $40.87\%$ for the strongest
baseline. It remains competitive across all five capabilities, with strong
results in coding and instruction following. 
Together with the LoRA/OFT results, these experiments show that the effectiveness of \methodname extends from parameter-efficient experts to fully fine-tuned
language experts.

\begin{wrapfigure}{r}{0.38\textwidth}
  \centering
  \vspace{-1em}
  \includegraphics[width=0.8\linewidth]{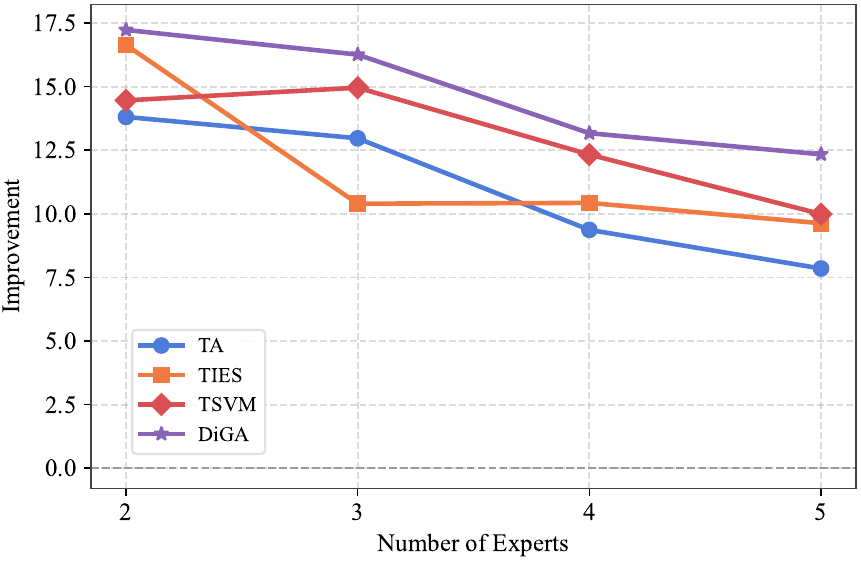}
  \vspace{-1em}
  \caption{Performance gains over the base model with different numbers of OFT experts.}
  \label{fig:geshu}
\end{wrapfigure}
\noindent\textbf{Vision-language expert merging.}
We further evaluate \methodname with three Qwen2.5-VL-7B-Instruct experts
specialized in OCR, spatial reasoning, and medical multimodal reasoning.
As shown in Table~\ref{tab:mm_results}, \methodname reaches an average score
of $61.84\%$, exceeding the base model by $1.52$ points and the strongest
compared baseline by $0.18$ points. It also approaches the individual-expert
average, with a gap of $0.23$ points. \methodname achieves the highest
PathVQA score of $67.61\%$ and remains competitive on the other four
benchmarks. These results show that the component-wise geometric treatment
also remains effective in vision-language expert merging.

\textbf{Scaling with the number of experts.}
Figure~\ref{fig:geshu} compares performance gains over the base model when merging two to five OFT experts. 
\methodname achieves the largest gain at each expert count, with the largest advantage observed in the five-expert setting. 
These results show that the benefit of \methodname persists as the number of merged experts increases from two to five within this expert suite.
Detailed results are reported in Appendix Tables~\ref{tab:2merge}, \ref{tab:3merge}, \ref{tab:4merge}, and~\ref{tab:5merge}.

\subsection{Ablation Studies}
\label{sec:ablations}

Table~\ref{tab:ablation} examines the contributions of geometric disentanglement and the component-wise aggregation design in the five-expert OFT setting.

\begin{table*}[t]
\centering
\caption{
\textbf{Generality and update-level effects of component-wise processing.}
Values in parentheses are changes relative to direct merging.
Component-Separate applies the base merger independently to the orientation and structural families, while Component-Joint applies it jointly to all decomposed components. Update statistics are
computed over the target matrices; per-task results are in Appendix Table~\ref{tab:tr_full_results1}.
$\Delta_{\mathrm{Comp}}$ denotes the merged update under each
component-processing mode. Norm ratio denotes
$\|\Delta_{\mathrm{Comp}}\|_F/\|\Delta_{\mathrm{Direct}}\|_F$;
relative distance denotes
$\|\Delta_{\mathrm{Comp}}-\Delta_{\mathrm{Direct}}\|_F/
\|\Delta_{\mathrm{Direct}}\|_F$; and zero fraction is the fraction of zero entries in $\Delta=W_{\mathrm{merged}}-W_{\mathrm{base}}$.
}
\label{tab:component_generality}

\resizebox{\textwidth}{!}{
\begin{tabular}{@{}llccrccc@{}}
\toprule
& & \multicolumn{2}{c}{\textbf{Performance}}
& \multicolumn{4}{c}{\textbf{Merged-update statistics}} \\
\cmidrule(lr){3-4}
\cmidrule(lr){5-8}

Base merger & Processing & Task Avg. & Transfer Avg.
& Norm ratio & \makecell{Zero frac.\\(Direct / Comp.)}
& Rel. dist. & Cosine \\
\midrule

TIES
& Component-Separate
& 56.08 {\scriptsize $(+0.26)$}
& 40.03 {\scriptsize $(+0.12)$}
& 0.989 & 0.0004 / 0.0278 & 0.592 & 0.823 \\

DARE
& Component-Separate
& 55.11 {\scriptsize $(+0.56)$}
& 36.66 {\scriptsize $(+0.13)$}
& 1.000 & 0.0232 / 0.0143 & 0.736 & 0.729 \\

TSVM
& Component-Separate
& 56.55 {\scriptsize $(+0.37)$}
& 40.02 {\scriptsize $(+0.44)$}
& 0.975 & 0.0203 / 0.0207 & 0.971 & 0.517 \\

TSVM
& Component-Joint
& 53.93 {\scriptsize $(-2.25)$}
& 40.17 {\scriptsize $(+0.59)$}
& 0.636 & 0.0203 / 0.0314 & 0.907 & 0.458 \\

\bottomrule
\end{tabular}
}

\end{table*}

\textbf{Contribution of the two geometric components.}
Retaining only the orientation component ($T$ only) reduces Task Avg. from
$58.60$ to $51.90$, whereas retaining only the structural component
($R$ only) yields $55.56$, corresponding to drops of $6.70$ and $3.04$
points, respectively. These results show that both components contribute
complementary task information, with the structural component accounting
for a larger share of the retained performance.

\textbf{Effect of component separation.}
We test the cross-component coupling hypothesis with
\textit{Whole-vector aggregation}, which uses the same aggregation rule as
\methodname but applies it to the complete task updates rather than
separating the orientation and structural families. Whole-vector aggregation
reduces Task Avg. from $58.60$ to $58.24$ and Transfer Avg. from $40.92$ to
$40.86$. Although modest, the consistent degradation supports the benefit
of preserving component identity during aggregation.

\textbf{Aggregation design.}
Removing redundancy-aware orthogonalization reduces Task Avg. by $0.39$
points, while removing consensus-aware magnitude calibration decreases it
by $0.33$ points. Removing source-magnitude weighting causes a larger
$3.42$-point drop, indicating that preserving component magnitudes is
particularly important for constructing the aggregate direction.
Replacing sequential Gram--Schmidt with order-independent symmetric
orthogonalization yields $58.23$, close to the full method. Together,
these results support the proposed aggregation design.

\subsection{Component Separation as a General Merging Principle}
\label{sec:sr_generality}

\textbf{Generality across existing merging operators.}
We examine whether the merging principle of \methodname, separating each task update into orientation and structural components and merging the two families independently---can also benefit existing methods. We apply this principle to TIES, DARE, and TSVM by replacing
$\mathcal{F}(\{\Delta_i\}_{i=1}^{N})$ with
$\mathcal{F}(\{T_i\}_{i=1}^{N})+\mathcal{F}(\{R_i\}_{i=1}^{N})$,
while otherwise preserving the original merging procedure.
As shown in Table~\ref{tab:component_generality}, component separation improves both Task Avg. and Transfer Avg. for all three methods.
The consistent improvements demonstrate that the idea of \methodname can be widely applied to other merging methods, 
highlighting its generality.

\textbf{Preserving component identity is essential.}
We test whether decomposition alone is sufficient or whether the two component families must remain separated during aggregation.
For TSVM, we compare \textit{Component-Joint}, which jointly aggregates all decomposed components, with \textit{Component-Separate}, which aggregates the two
families independently before recombination. Component-Separate improves Task Avg. by $0.37$ points over TSVM, whereas Component-Joint decreases it
by $2.25$ points. Their $2.62$-point gap shows that decomposition alone is insufficient; preserving component identity during aggregation is essential
to avoid cross-component coupling.

\textbf{Geometric characteristics of the merged update.}
We further examine how component separation alters the merged update.
As shown in Table~\ref{tab:component_generality}, the norm ratios remain close to one, while the component-separated solutions differ substantially in direction from their direct-merging counterparts.
Thus, the gains cannot be explained by increased adaptation magnitude; component separation primarily changes the direction of the merged update, consistent with the proposed effect of avoiding cross-component coupling during aggregation.

\begin{figure*}[t]
\centering
\includegraphics[width=\linewidth]{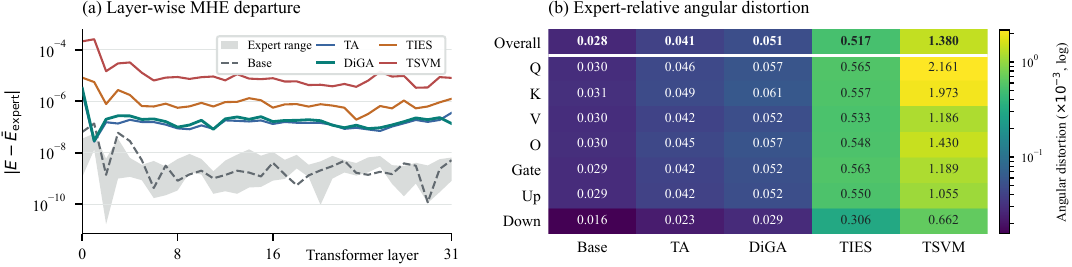}
\vspace{-1.8em}
\caption{\textbf{Geometric effects of merging on OFT experts.}
(a) Layer-wise MHE departure from the expert regime.
(b) Expert-relative angular distortion overall and across projection types.}
\label{fig:geometric_effects}
\end{figure*}

\subsection{Geometric Effects of Merging}
\label{sec:geometric_effects}

We examine how different merging strategies reshape the internal directional
geometry of the resulting weights. Following Minimum Hyperspherical Energy
(MHE)~\citep{NEURIPS2018_177540c7}, we treat normalized row vectors of each
attention and MLP projection matrix as points on a unit hypersphere. For each
matrix, we compute its hyperspherical energy and report the layer-wise
absolute deviation from the mean energy of the five OFT experts.
Figure~\ref{fig:geometric_effects}(a) shows that the experts occupy a narrow
geometric regime, while merging methods depart from it to different extents.

We further compare the pairwise row-wise cosine structure of each merged
matrix with that of the individual experts and average the absolute
differences across experts and matrices.
Figure~\ref{fig:geometric_effects}(b) shows that closer alignment with expert
geometry does not necessarily yield better merging performance. The base
model and TA remain closer to the expert angular structures than \methodname
but achieve lower OFT Task Avg., whereas TIES and TSVM introduce much larger
distortions. \methodname lies between these extremes, achieving the strongest
OFT Task Avg. while avoiding excessive disruption of the expert-induced
directional structure. The same trend holds across attention and MLP
projections. Similar effects
are observed for LoRA experts in Appendix Figure~\ref{fig:geometric_effects_lora}.

These results suggest that effective merging does not simply preserve expert geometry as closely as possible. Rather, successful merging appears to require a controlled reorganization of the underlying directional structure, balancing geometric preservation with the flexibility needed to integrate multiple expert capabilities. More broadly, this suggests a promising direction for future model merging: designing objectives that distinguish geometry that should be preserved from geometry that can be adaptively reorganized, rather than treating geometric similarity itself as the optimization target.

\section{Related Work}

\textbf{Structure and Geometry of Model Adaptation.}
Prior work shows that downstream adaptation exhibits structured parameter changes, often within low-dimensional subspaces, motivating methods such as LoRA~\citep{hu2022lora,liu2026smoa}.
Orthogonal fine-tuning models adaptation as rotations of pretrained weights that preserve pairwise angular relationships~\citep{qiu2023oft,liu2024parameter}, while qGOFT relaxes this constraint to allow controlled changes in norms and angles~\citep{ma2024parameter}. Building on these observations, we ask whether such directional-reorganization geometry can be isolated within general task updates and exploited for model merging.

\textbf{Model Merging.}
Model merging combines independently adapted models without joint retraining, from weight averaging~\citep{wang2026dim,wang2026more,wang2026plam} and task arithmetic~\citep{ilharco2023taskarithmetic} to methods that mitigate interference through sparsification, sign resolution, parameter statistics, or adaptive weighting~\citep{yadav2023ties,yu2024dare,jin2023dataless,yang2024adamerging}. Recent approaches further exploit structured task representations: 
TSVM~\citep{gargiulo2025tsv} uses singular-vector representations with orthogonalization, while Iso-CTS~\citep{marczak2025no} models common and task-specific subspaces.
In contrast, our work separates geometrically distinct components within each task update before aggregation.

\section{Conclusion}
We identify \emph{cross-component coupling} as a previously overlooked source
of interference in model merging and propose \textbf{DiGA}, which disentangles
task updates into orthogonal orientation and structural components and merges
them separately before recombination. Experiments across language and
vision-language experts show that DiGA improves merged-model performance and
reduces capability degradation. Results with existing merging methods further
support component-wise processing as a general and reusable merging principle.

\section*{AI Use Statement}

In this work, we used generative AI tools for improving the readability and language quality of the manuscript. All AI-assisted modifications were reviewed and verified by the authors. We take full responsibility for the final content of this work, including all text, claims, mathematical formulations, and
experimental results.

\section*{Reproducibility Statement}

We provide an anonymous repository containing the implementation of
\methodname, together with the merging algorithms, experimental scripts,
and evaluation code used in this work. Detailed experimental settings,
datasets, baselines, and implementation configurations are documented in
Section~\ref{sec:experimental_setup} and the appendix. The appendix also
provides the assumptions, derivations, and proofs for the base-induced
orthogonal decomposition, cross-component orthogonality, and
cross-component coupling analysis.

\bibliography{iclr2027_conference}

@inproceedings{wang2026more,
  title={Why do more experts fail? a theoretical analysis of model merging},
  author={Wang, Zijing and Xu, Xingle and Liu, Yongkang and Zhang, Yiqun and Lin, Peiqin and Feng, Shi and Wang, Daling and Yang, Xiaocui and Sch{\"u}tze, Hinrich},
  booktitle={Proceedings of the 64th Annual Meeting of the Association for Computational Linguistics (Volume 1: Long Papers)},
  pages={45460--45482},
  year={2026}
}

@inproceedings{yue2024mmmu,
  title={Mmmu: A massive multi-discipline multimodal understanding and reasoning benchmark for expert agi},
  author={Yue, Xiang and Ni, Yuansheng and Zhang, Kai and Zheng, Tianyu and Liu, Ruoqi and Zhang, Ge and Stevens, Samuel and Jiang, Dongfu and Ren, Weiming and Sun, Yuxuan and others},
  booktitle={Proceedings of the IEEE/CVF conference on computer vision and pattern recognition},
  pages={9556--9567},
  year={2024}
}

@article{yang2026model,
  title={Model merging in llms, mllms, and beyond: Methods, theories, applications, and opportunities},
  author={Yang, Enneng and Shen, Li and Guo, Guibing and Wang, Xingwei and Cao, Xiaochun and Zhang, Jie and Tao, Dacheng},
  journal={ACM Computing Surveys},
  volume={58},
  number={8},
  pages={1--41},
  year={2026},
  publisher={ACM New York, NY}
}

@article{zhou2023instruction,
  title={Instruction-following evaluation for large language models},
  author={Zhou, Jeffrey and Lu, Tianjian and Mishra, Swaroop and Brahma, Siddhartha and Basu, Sujoy and Luan, Yi and Zhou, Denny and Hou, Le},
  journal={arXiv preprint arXiv:2311.07911},
  year={2023}
}

@inproceedings{wei2026designing,
  title={Designing incident reporting systems for harms from general-purpose AI},
  author={Wei, Kevin and Heim, Lennart},
  booktitle={Proceedings of the AAAI Conference on Artificial Intelligence},
  volume={40},
  number={44},
  pages={38016--38029},
  year={2026}
}

@article{legg2007universal,
  title={Universal intelligence: A definition of machine intelligence},
  author={Legg, Shane and Hutter, Marcus},
  journal={Minds and machines},
  volume={17},
  number={4},
  pages={391--444},
  year={2007},
  publisher={Springer}
}

@article{bommasani2021opportunities,
  title={On the opportunities and risks of foundation models},
  author={Bommasani, Rishi and Hudson, Drew A and Adeli, Ehsan and Altman, Russ and Arora, Simran and von Arx, Sydney and Bernstein, Michael S and Bohg, Jeannette and Bosselut, Antoine and Brunskill, Emma and others},
  journal={arXiv preprint arXiv:2108.07258},
  year={2021}
}

@article{hoffmann2022training,
  title={Training compute-optimal large language models},
  author={Hoffmann, Jordan and Borgeaud, Sebastian and Mensch, Arthur and Buchatskaya, Elena and Cai, Trevor and Rutherford, Eliza and Casas, Diego de Las and Hendricks, Lisa Anne and Welbl, Johannes and Clark, Aidan and others},
  journal={arXiv preprint arXiv:2203.15556},
  year={2022}
}

@article{kaplan2020scaling,
  title={Scaling laws for neural language models},
  author={Kaplan, Jared and McCandlish, Sam and Henighan, Tom and Brown, Tom B and Chess, Benjamin and Child, Rewon and Gray, Scott and Radford, Alec and Wu, Jeffrey and Amodei, Dario},
  journal={arXiv preprint arXiv:2001.08361},
  year={2020}
}

@inproceedings{
li2026when,
title={When Shared Knowledge Hurts: Spectral Over-Accumulation in Model Merging},
author={Yayuan Li and Ze Peng and Jian Zhang and Jintao Guo and Yue Duan and Yinghuan Shi},
booktitle={Forty-third International Conference on Machine Learning},
year={2026},
url={https://openreview.net/forum?id=WUjk3RVZyV}
}

@inproceedings{
ma2024parameter,
title={Parameter Efficient Quasi-Orthogonal Fine-Tuning via Givens Rotation},
author={Xinyu Ma and Xu Chu and Zhibang Yang and Yang Lin and Xin Gao and Junfeng Zhao},
booktitle={Forty-first International Conference on Machine Learning},
year={2024},
url={https://openreview.net/forum?id=1zFkjbTgwC}
}

@inproceedings{NEURIPS2018_177540c7,
 author = {Liu, Weiyang and Lin, Rongmei and Liu, Zhen and Liu, Lixin and Yu, Zhiding and Dai, Bo and Song, Le},
 booktitle = {Advances in Neural Information Processing Systems},
 editor = {S. Bengio and H. Wallach and H. Larochelle and K. Grauman and N. Cesa-Bianchi and R. Garnett},
 pages = {},
 publisher = {Curran Associates, Inc.},
 title = {Learning towards Minimum Hyperspherical Energy},
 url = {https://proceedings.neurips.cc/paper_files/paper/2018/file/177540c7bcb8db31697b601642eac8d4-Paper.pdf},
 volume = {31},
 year = {2018}
}

@inproceedings{
chen2025bring,
title={Bring Reason to Vision: Understanding Perception and Reasoning through Model Merging},
author={Shiqi Chen and Jinghan Zhang and Tongyao Zhu and Wei Liu and Siyang Gao and Miao Xiong and Manling Li and Junxian He},
booktitle={Forty-second International Conference on Machine Learning},
year={2025},
url={https://openreview.net/forum?id=ntCAP6tMoX}
}

@article{cobbe2021training,
  title={Training verifiers to solve math word problems},
  author={Cobbe, Karl and Kosaraju, Vineet and Bavarian, Mohammad and Chen, Mark and Jun, Heewoo and Kaiser, Lukasz and Plappert, Matthias and Tworek, Jerry and Hilton, Jacob and Nakano, Reiichiro and others},
  journal={arXiv preprint arXiv:2110.14168},
  year={2021}
}

@article{he2026mergebench,
  title={Mergebench: A benchmark for merging domain-specialized llms},
  author={He, Yifei and Zeng, Siqi and Hu, Yuzheng and Yang, Rui and Zhang, Tong and Zhao, Han},
  journal={Advances in Neural Information Processing Systems},
  volume={38},
  year={2026}
}

@inproceedings{wei2024magicoder,
  title={Magicoder: Empowering Code Generation with OSS-Instruct},
  author={Wei, Yuxiang and Wang, Zhe and Liu, Jiawei and Ding, Yifeng and Zhang, Lingming},
  booktitle={International Conference on Machine Learning},
  pages={52632--52657},
  year={2024},
  organization={PMLR}
}

@article{clark2018think,
  title={Think you have solved question answering? try arc, the ai2 reasoning challenge},
  author={Clark, Peter and Cowhey, Isaac and Etzioni, Oren and Khot, Tushar and Sabharwal, Ashish and Schoenick, Carissa and Tafjord, Oyvind},
  journal={arXiv preprint arXiv:1803.05457},
  year={2018}
}

@inproceedings{cai2026scaling,
  title={Scaling spatial intelligence with multimodal foundation models},
  author={Cai, Zhongang and Wang, Ruisi and Gu, Chenyang and Pu, Fanyi and Xu, Junxiang and Wang, Yubo and Yin, Wanqi and Yang, Zhitao and Wei, Chen and Zhou, Tongxi and others},
  booktitle={Proceedings of the IEEE/CVF Conference on Computer Vision and Pattern Recognition},
  pages={7879--7890},
  year={2026}
}

@article{poznanski2025olmocr,
  title={olmocr 2: Unit test rewards for document ocr},
  author={Poznanski, Jake and Soldaini, Luca and Lo, Kyle},
  journal={arXiv preprint arXiv:2510.19817},
  year={2025}
}

@article{chen2406huatuogptvision,
  title={Huatuogptvision, towards injecting medical visual knowledge into multimodal llms at scale. CoRR, abs/2406.19280, 2024b. doi: 10.48550},
  author={Chen, Junying and Ouyang, Ruyi and Gao, Anningzhe and Chen, Shunian and Chen, Guiming Hardy and Wang, Xidong and Zhang, Ruifei and Cai, Zhenyang and Ji, Ke and Yu, Guangjun and others},
  journal={arXiv preprint ARXIV.2406.19280},
  year={2024}
}

@article{yang2025mmsi,
  title={Mmsi-bench: A benchmark for multi-image spatial intelligence},
  author={Yang, Sihan and Xu, Runsen and Xie, Yiman and Yang, Sizhe and Li, Mo and Lin, Jingli and Zhu, Chenming and Chen, Xiaochen and Duan, Haodong and Yue, Xiangyu and others},
  journal={arXiv preprint arXiv:2505.23764},
  year={2025}
}

@inproceedings{du2024embspatial,
  title={Embspatial-bench: Benchmarking spatial understanding for embodied tasks with large vision-language models},
  author={Du, Mengfei and Wu, Binhao and Li, Zejun and Huang, Xuan-Jing and Wei, Zhongyu},
  booktitle={Proceedings of the 62nd Annual Meeting of the Association for Computational Linguistics (Volume 2: Short Papers)},
  pages={346--355},
  year={2024}
}

@article{liu2024ocrbench,
  title={Ocrbench: on the hidden mystery of ocr in large multimodal models},
  author={Liu, Yuliang and Li, Zhang and Huang, Mingxin and Yang, Biao and Yu, Wenwen and Li, Chunyuan and Yin, Xu-Cheng and Liu, Cheng-Lin and Jin, Lianwen and Bai, Xiang},
  journal={Science China Information Sciences},
  volume={67},
  number={12},
  pages={220102},
  year={2024},
  publisher={Springer}
}

@article{he2020pathvqa,
  title={Pathvqa: 30000+ questions for medical visual question answering},
  author={He, Xuehai and Zhang, Yichen and Mou, Luntian and Xing, Eric and Xie, Pengtao},
  journal={arXiv preprint arXiv:2003.10286},
  year={2020}
}

@inproceedings{zhong2024agieval,
  title={Agieval: A human-centric benchmark for evaluating foundation models},
  author={Zhong, Wanjun and Cui, Ruixiang and Guo, Yiduo and Liang, Yaobo and Lu, Shuai and Wang, Yanlin and Saied, Amin and Chen, Weizhu and Duan, Nan},
  booktitle={Findings of the association for computational linguistics: NAACL 2024},
  pages={2299--2314},
  year={2024}
}

@article{li2024numinamath,
  title={Numinamath: The largest public dataset in ai4maths with 860k pairs of competition math problems and solutions},
  author={Li, Jia and Beeching, Edward and Tunstall, Lewis and Lipkin, Ben and Soletskyi, Roman and Huang, Shengyi and Rasul, Kashif and Yu, Longhui and Jiang, Albert Q and Shen, Ziju and others},
  journal={Hugging Face repository},
  volume={13},
  number={9},
  pages={9},
  year={2024}
}

@article{liu2023your,
  title={Is your code generated by chatgpt really correct? rigorous evaluation of large language models for code generation},
  author={Liu, Jiawei and Xia, Chunqiu Steven and Wang, Yuyao and Zhang, Lingming},
  journal={Advances in neural information processing systems},
  volume={36},
  pages={21558--21572},
  year={2023}
}

@inproceedings{hendrycks2measuring,
  title={Measuring Mathematical Problem Solving With the MATH Dataset},
  author={Hendrycks, Dan and Burns, Collin and Kadavath, Saurav and Arora, Akul and Basart, Steven and Tang, Eric and Song, Dawn and Steinhardt, Jacob},
  booktitle={Thirty-fifth Conference on Neural Information Processing Systems Datasets and Benchmarks Track (Round 2)}
}

@inproceedings{sap2019social,
  title={Social IQa: Commonsense reasoning about social interactions},
  author={Sap, Maarten and Rashkin, Hannah and Chen, Derek and Le Bras, Ronan and Choi, Yejin},
  booktitle={Proceedings of the 2019 conference on empirical methods in natural language processing and the 9th international joint conference on natural language processing (EMNLP-IJCNLP)},
  pages={4463--4473},
  year={2019}
}

@inproceedings{talmor2019commonsenseqa,
  title={Commonsenseqa: A question answering challenge targeting commonsense knowledge},
  author={Talmor, Alon and Herzig, Jonathan and Lourie, Nicholas and Berant, Jonathan},
  booktitle={Proceedings of the 2019 Conference of the North American Chapter of the Association for Computational Linguistics: Human Language Technologies, Volume 1 (Long and Short Papers)},
  pages={4149--4158},
  year={2019}
}

@article{lu2022learn,
  title={Learn to explain: Multimodal reasoning via thought chains for science question answering},
  author={Lu, Pan and Mishra, Swaroop and Xia, Tanglin and Qiu, Liang and Chang, Kai-Wei and Zhu, Song-Chun and Tafjord, Oyvind and Clark, Peter and Kalyan, Ashwin},
  journal={Advances in neural information processing systems},
  volume={35},
  pages={2507--2521},
  year={2022}
}

@article{grattafiori2024llama,
  title={The Llama 3 Herd of Models},
  author={Grattafiori, Aaron and Dubey, Abhimanyu and Jauhri, Abhinav and others},
  journal={arXiv preprint arXiv:2407.21783},
  year={2024}
}

@article{bai2025qwen25vl,
  title={Qwen2.5-VL Technical Report},
  author={Bai, Shuai and Chen, Keqin and Liu, Xuejing and Wang, Jialin and Ge, Wenbin and Song, Sibo and Dang, Kai and Wang, Peng and Wang, Shijie and Tang, Jun and others},
  journal={arXiv preprint arXiv:2502.13923},
  year={2025}
}

@inproceedings{ilharco2023taskarithmetic,
  title={Editing Models with Task Arithmetic},
  author={Ilharco, Gabriel and Ribeiro, Marco Tulio and Wortsman, Mitchell and Gururangan, Suchin and Schmidt, Ludwig and Hajishirzi, Hannaneh and Farhadi, Ali},
  booktitle={International Conference on Learning Representations},
  year={2023}
}

@inproceedings{yadav2023ties,
  title={TIES-Merging: Resolving Interference When Merging Models},
  author={Yadav, Prateek and Tam, Derek and Choshen, Leshem and Raffel, Colin and Bansal, Mohit},
  booktitle={Advances in Neural Information Processing Systems},
  year={2023}
}

@inproceedings{yu2024dare,
  title={Language Models are Super Mario: Absorbing Abilities from Homologous Models as a Free Lunch},
  author={Yu, Le and Yu, Bowen and Yu, Haiyang and Huang, Fei and Li, Yongbin},
  booktitle={International Conference on Machine Learning},
  year={2024}
}

@inproceedings{gargiulo2025tsv,
  title={Task singular vectors: Reducing task interference in model merging},
  author={Gargiulo, Antonio Andrea and Crisostomi, Donato and Bucarelli, Maria Sofia and Scardapane, Simone and Silvestri, Fabrizio and Rodola, Emanuele},
  booktitle={Proceedings of the Computer Vision and Pattern Recognition Conference},
  pages={18695--18705},
  year={2025}
}

@article{yang2026orthomerge,
  title={Orthogonal Model Merging},
  author={Yang, Sihan and Shi, Kexuan and Liu, Weiyang},
  journal={arXiv preprint arXiv:2602.05943},
  year={2026}
}

@inproceedings{hu2022lora,
  title={LoRA: Low-Rank Adaptation of Large Language Models},
  author={Hu, Edward J. and Shen, Yelong and Wallis, Phillip and Allen-Zhu, Zeyuan and Li, Yuanzhi and Wang, Shean and Wang, Lu and Chen, Weizhu},
  booktitle={International Conference on Learning Representations},
  year={2022}
}

@article{qiu2023oft,
  title={Controlling Text-to-Image Diffusion by Orthogonal Finetuning},
  author={Qiu, Zeju and Liu, Weiyang and Feng, Haiwen and Xue, Yuxuan and Feng, Yujun and Liu, Zhen and Zhang, Dan and Weller, Adrian and Sch{\"o}lkopf, Bernhard},
  journal={Advances in Neural Information Processing Systems},
  volume={36},
  pages={79320--79362},
  year={2023}
}

@article{wang2025scaling,
  title={Scaling Intelligence Through Model Merging: A Comprehensive Survey},
  author={Wang, Zijing and Liu, Yongkang and Luo, Yingfeng and Wang, Ming and Song, Zhen and Feng, Shi and Yang, Xiaocui and Lin, Dingyang and Wang, Daling and Zhang, Yifei and others},
  year={2025},
  publisher={TechRxiv}
}

@inproceedings{wang2026plam,
  title={PlaM: Training-Free Plateau-Guided Model Merging for Better Visual Grounding in MLLMs},
  author={Wang, Zijing and Liu, Yongkang and Wang, Mingyang and Nie, Ercong and Chen, Deyuan and Zhao, Zhengjie and Feng, Shi and Wang, Daling and Yang, Xiaocui and Zhang, Yifei and others},
  booktitle={Findings of the Association for Computational Linguistics: ACL 2026},
  pages={21019--21035},
  year={2026}
}

@article{bjorck1967solving,
  title={Solving linear least squares problems by Gram-Schmidt orthogonalization},
  author={Bj{\"o}rck, {\AA}ke},
  journal={BIT Numerical Mathematics},
  volume={7},
  number={1},
  pages={1--21},
  year={1967},
  publisher={Springer}
}

@inproceedings{
marczak2025no,
title={No Task Left Behind: Isotropic Model Merging with Common and Task-Specific Subspaces},
author={Daniel Marczak and Simone Magistri and Sebastian Cygert and Bart{\l}omiej Twardowski and Andrew D. Bagdanov and Joost van de Weijer},
booktitle={Forty-second International Conference on Machine Learning},
year={2025},
url={https://openreview.net/forum?id=RBZpAa27ls}
}

@inproceedings{mitchellwortsman2022model,
  title={Model soups: averaging weights of multiple fine-tuned models improves accuracy without increasing inference time},
  author={MitchellWortsman, Gabriel Ilharco},
  booktitle={Proceedings of the 39th International Conference on Machine Learning, Baltimore, Maryland},
  year={2022}
}

@article{huang2024emr,
  title={Emr-merging: Tuning-free high-performance model merging},
  author={Huang, Chenyu and Ye, Peng and Chen, Tao and He, Tong and Yue, Xiangyu and Ouyang, Wanli},
  journal={Advances in Neural Information Processing Systems},
  volume={37},
  pages={122741--122769},
  year={2024}
}

@inproceedings{
jin2023dataless,
title={Dataless Knowledge Fusion by Merging Weights of Language Models},
author={Xisen Jin and Xiang Ren and Daniel Preotiuc-Pietro and Pengxiang Cheng},
booktitle={The Eleventh International Conference on Learning Representations },
year={2023},
url={https://openreview.net/forum?id=FCnohuR6AnM}
}

@article{wang2026dim,
  title={DiM$\backslash$textsuperscript $\{$3$\}$: Bridging Multilingual and Multimodal Models via Direction-and Magnitude-Aware Merging},
  author={Wang, Zijing and Wang, Mingyang and Nie, Ercong and Liu, Yongkang and Feng, Shi and Zhao, Mengjie and Wang, Daling and Yang, Xiaocui and Sch{\"u}tze, Hinrich},
  journal={arXiv preprint arXiv:2605.12960},
  year={2026}
}

@inproceedings{yang2024adamerging,
  title={Adamerging: Adaptive model merging for multi-task learning},
  author={Yang, Enneng and Wang, Zhenyi and Shen, Li and Liu, Shiwei and Guo, Guibing and Wang, Xingwei and Tao, Dacheng},
  booktitle={International Conference on Learning Representations},
  volume={2024},
  pages={22743--22763},
  year={2024}
}

@inproceedings{lee2025star,
  title={Star: Spectral truncation and rescale for model merging},
  author={Lee, Yu-Ang and Ko, Ching-Yun and Pedapati, Tejaswini and Chung, I-Hsin and Yeh, Mi-Yen and Chen, Pin-Yu},
  booktitle={Proceedings of the 2025 Conference of the Nations of the Americas Chapter of the Association for Computational Linguistics: Human Language Technologies (Volume 2: Short Papers)},
  pages={496--505},
  year={2025}
}

@inproceedings{stoica2025model,
  title={Model merging with svd to tie the knots},
  author={Stoica, George and Ramesh, Pratik and Ecsedi, Boglarka and Choshen, Leshem and Hoffman, Judy},
  booktitle={International Conference on Learning Representations},
  volume={2025},
  pages={4501--4519},
  year={2025}
}

@article{liu2026smoa,
  title={SMoA: Spectrum Modulation Adapter for Parameter-Efficient Fine-Tuning},
  author={Liu, Yongkang and Li, Xing and Zhao, Mengjie and Zhang, Shanru and Wang, Zijing and Li, Qian and Feng, Shi and Ren, Feiliang and Wang, Daling and Sch{\"u}tze, Hinrich},
  journal={arXiv preprint arXiv:2605.21147},
  year={2026}
}

@inproceedings{liu2024parameter,
  title={Parameter-efficient orthogonal finetuning via butterfly factorization},
  author={Liu, Weiyang and Qiu, Zeju and Feng, Yao and Xiu, Yuliang and Xue, Yuxuan and Yu, Longhui and Feng, Haiwen and Liu, Zhen and Heo, Juyeon and Peng, Songyou and others},
  booktitle={International Conference on Learning Representations},
  volume={2024},
  pages={38317--38350},
  year={2024}
}

@article{austin2021program,
  title={Program synthesis with large language models},
  author={Austin, Jacob and Odena, Augustus and Nye, Maxwell and Bosma, Maarten and Michalewski, Henryk and Dohan, David and Jiang, Ellen and Cai, Carrie and Terry, Michael and Le, Quoc and others},
  journal={arXiv preprint arXiv:2108.07732},
  year={2021}
}

@article{han2024wildguard,
  title={Wildguard: Open one-stop moderation tools for safety risks, jailbreaks, and refusals of llms},
  author={Han, Seungju and Rao, Kavel and Ettinger, Allyson and Jiang, Liwei and Lin, Bill Yuchen and Lambert, Nathan and Choi, Yejin and Dziri, Nouha},
  journal={Advances in neural information processing systems},
  volume={37},
  pages={8093--8131},
  year={2024}
}

@InProceedings{pmlr-v235-mazeika24a,
  title = 	 {{H}arm{B}ench: A Standardized Evaluation Framework for Automated Red Teaming and Robust Refusal},
  author =       {Mazeika, Mantas and Phan, Long and Yin, Xuwang and Zou, Andy and Wang, Zifan and Mu, Norman and Sakhaee, Elham and Li, Nathaniel and Basart, Steven and Li, Bo and Forsyth, David and Hendrycks, Dan},
  booktitle = 	 {Proceedings of the 41st International Conference on Machine Learning},
  pages = 	 {35181--35224},
  year = 	 {2024},
  editor = 	 {Salakhutdinov, Ruslan and Kolter, Zico and Heller, Katherine and Weller, Adrian and Oliver, Nuria and Scarlett, Jonathan and Berkenkamp, Felix},
  volume = 	 {235},
  series = 	 {Proceedings of Machine Learning Research},
  month = 	 {21--27 Jul},
  publisher =    {PMLR},
  url = 	 {https://proceedings.mlr.press/v235/mazeika24a.html}
}

@inproceedings{rottger2024xstest,
  title={Xstest: A test suite for identifying exaggerated safety behaviours in large language models},
  author={R{\"o}ttger, Paul and Kirk, Hannah and Vidgen, Bertie and Attanasio, Giuseppe and Bianchi, Federico and Hovy, Dirk},
  booktitle={Proceedings of the 2024 Conference of the North American Chapter of the Association for Computational Linguistics: Human Language Technologies (Volume 1: Long Papers)},
  pages={5377--5400},
  year={2024}
}

@inproceedings{shen2024anything,
  title={" do anything now": Characterizing and evaluating in-the-wild jailbreak prompts on large language models},
  author={Shen, Xinyue and Chen, Zeyuan and Backes, Michael and Shen, Yun and Zhang, Yang},
  booktitle={Proceedings of the 2024 on ACM SIGSAC Conference on Computer and Communications Security},
  pages={1671--1685},
  year={2024}
}

@article{biderman2024lessons,
  title={Lessons from the trenches on reproducible evaluation of language models},
  author={Biderman, Stella and Schoelkopf, Hailey and Sutawika, Lintang and Gao, Leo and Tow, Jonathan and Abbasi, Baber and Aji, Alham Fikri and Ammanamanchi, Pawan Sasanka and Black, Sidney and Clive, Jordan and others},
  journal={arXiv preprint arXiv:2405.14782},
  year={2024}
}

@inproceedings{zhang2025lmms,
  title={Lmms-eval: Reality check on the evaluation of large multimodal models},
  author={Zhang, Kaichen and Li, Bo and Zhang, Peiyuan and Pu, Fanyi and Cahyono, Joshua Adrian and Hu, Kairui and Liu, Shuai and Zhang, Yuanhan and Yang, Jingkang and Li, Chunyuan and others},
  booktitle={Findings of the Association for Computational Linguistics: NAACL 2025},
  pages={881--916},
  year={2025}
}

@misc{allal2022framework,
  title={A framework for the evaluation of code generation models},
  author={Allal, Loubna Ben and Muennighoff, Niklas and Umapathi, Logesh Kumar and Lipkin, Ben and Von Werra, Leandro},
  year={2022}
}

@inproceedings{lightman2024let,
  title={Let's verify step by step},
  author={Lightman, Hunter and Kosaraju, Vineet and Burda, Yuri and Edwards, Harrison and Baker, Bowen and Lee, Teddy and Leike, Jan and Schulman, John and Sutskever, Ilya and Cobbe, Karl},
  booktitle={International Conference on Learning Representations},
  volume={2024},
  pages={39578--39601},
  year={2024}
}
\bibliographystyle{iclr2027_conference}

\newpage
\appendix

\section{Experimental Setup}
\label{app:experimental_setup}

We compare \methodname with representative data-free merging methods, including Task Arithmetic (TA)~\citep{ilharco2023taskarithmetic}, TIES-Merging~\citep{yadav2023ties}, DARE~\citep{yu2024dare}, Task Singular Vector Merging (TSVM)~\citep{gargiulo2025tsv}, and the conflict-aware and global variants of OrthoMerge~\citep{yang2026orthomerge}, denoted OrthoMerge-C and OrthoMerge-G. Baseline merging coefficients follow the evaluation protocol of OrthoMerge and are fixed before test evaluation, with no tuning on the test sets. 
For \methodname, we apply the proposed decomposition to the two-dimensional attention and MLP projection matrices in each Transformer block. Parameters not covered by this matrix decomposition, such as embeddings, normalization parameters, and the language-model head, are merged using TA. Decoding and evaluation follow the standard protocol of each benchmark.

We evaluate \methodname across language-only and vision-language models,
covering expert models obtained through parameter-efficient and full
fine-tuning and spanning diverse specialized capabilities. Our experiments
are designed to assess not only the overall merging performance of
\methodname, but also whether its gains arise from the proposed
component-wise geometric disentanglement and whether this principle
generalizes across different merging operators.

\subsection{Language-only experts}

\paragraph{Expert based on LoRA fine-tuning}
We first evaluate \methodname on language-only experts. For
Llama-3.1-8B~\citep{grattafiori2024llama}, we use two matched suites of
expert checkpoints from OrthoMerge~\citep{yang2026orthomerge}, adapted with
LoRA and Orthogonal Finetuning (OFT), respectively. Each suite contains five
experts specialized in code generation, mathematical reasoning, scientific
question answering, commonsense reasoning, and social knowledge, trained on
Magicoder-OSS-Instruct~\citep{wei2024magicoder},
NuminaMath-TIR~\citep{li2024numinamath},
ScienceQA~\citep{lu2022learn},
CommonsenseQA~\citep{talmor2019commonsenseqa}, and
Social-IQA~\citep{sap2019social}, respectively.
We evaluate these capabilities on HumanEval+~\citep{liu2023your},
MATH500~\citep{lightman2024let}, ScienceQA, CommonsenseQA, and Social-IQA. To assess transfer beyond the training
domains, 
we also evaluate on AGIEval~\citep{zhong2024agieval}
and ARC-Avg, the average accuracy over 12 machine-translated ARC-Challenge
subsets~\citep{clark2018think}. We report the macro averages over the five
target-task benchmarks and the two transfer benchmarks as \textit{Task Avg.}
and \textit{Transfer Avg.}, respectively.

\paragraph{Expert based on full fine-tuning}
We evaluate fully fine-tuned experts from MergeBench~\citep{he2026mergebench} by merging five Llama-3.2-3B~\citep{grattafiori2024llama} experts specialized in instruction following, mathematics, coding, multilinguality, and safety. Instruction following is evaluated on IFEval~\citep{zhou2023instruction}, mathematics on GSM8K~\citep{cobbe2021training} with chain-of-thought prompting, coding on HumanEval+~\citep{liu2023your} and MBPP+~\citep{austin2021program}, and multilinguality on ARC-Avg. Safety is evaluated on WildGuardTest~\citep{han2024wildguard}, HarmBench~\citep{pmlr-v235-mazeika24a}, XSTest~\citep{rottger2024xstest}, and DoAnythingNow~\citep{shen2024anything}. For capabilities with multiple benchmarks, we first average their benchmark scores and then compute the macro average across the five capabilities.

\subsection{Vision-language experts}
We further evaluate the proposed merging principle in a representative vision-language setting. Using Qwen2.5-VL-7B-Instruct~\citep{bai2025qwen25vl} as the shared base model, we merge three experts specialized in spatial reasoning, optical character recognition, and medical multimodal reasoning: SenseNova-SI-1.1-Qwen2.5-VL-7B~\citep{cai2026scaling}, olmOCR-2-7B-1025~\citep{poznanski2025olmocr}, and HuatuoGPT-Vision-7B~\citep{chen2406huatuogptvision}, respectively. Spatial reasoning is evaluated on MMSI-Bench~\citep{yang2025mmsi} and EmbSpatial~\citep{du2024embspatial}, optical character recognition on OCRBench~\citep{liu2024ocrbench}, and medical multimodal reasoning on MMMU-Med~\citep{yue2024mmmu} and PathVQA~\citep{he2020pathvqa}. We report the macro average across the five benchmarks.

\section{Benchmarks}
\label{app:bench}

\textbf{MATH500}~\citep{lightman2024let} is a mathematical reasoning benchmark derived from the MATH~\citep{hendrycks2measuring} dataset. It contains 500 challenging competition-level mathematics problems spanning seven subjects and evaluates models' mathematical reasoning and problem-solving abilities.

\textbf{HumanEval+}~\citep{liu2023your} extends HumanEval with substantially more test cases for each programming problem, providing a stricter assessment of the functional correctness of model-generated code.

\textbf{ScienceQA}~\citep{lu2022learn} is a large-scale multiple-choice benchmark covering natural, social, and language science. We evaluate on image-free test examples using only the questions and answer choices, excluding hints, lectures, and explanations, and report answer accuracy.

\textbf{CommonsenseQA}~\citep{talmor2019commonsenseqa} is a multiple-choice question answering benchmark designed to evaluate commonsense knowledge and reasoning. It requires models to infer implicit relationships and leverage prior knowledge to answer questions.

\textbf{Social-IQA}~\citep{sap2019social} is a multiple-choice benchmark for social commonsense reasoning. It evaluates whether models can infer people's intentions, reactions, and motivations in everyday social situations.

\textbf{AGIEval}~\citep{zhong2024agieval} is a bilingual, human-centric benchmark comprising questions from official admission and qualification examinations, as well as academic competitions, in both Chinese and English. We use the full benchmark to evaluate the out-of-domain generalization of merged models across diverse exam domains and both languages.

\textbf{ARC-Avg} denotes the average accuracy over 12 machine-translated variants of ARC-Challenge, the difficult subset of the AI2 Reasoning Challenge (ARC)~\citep{clark2018think}, which comprises grade-school multiple-choice science questions requiring scientific knowledge and reasoning.

\textbf{MBPP+} extends MBPP~\citep{austin2021program} with substantially expanded test cases for 399 curated tasks, enabling a more rigorous execution-based evaluation of functional correctness.

\textbf{GSM8K}~\citep{cobbe2021training} is a dataset of 8.5K linguistically diverse grade-school mathematics word problems that require multi-step reasoning using elementary arithmetic. We evaluate on its test set using 8-shot chain-of-thought prompting and final-answer exact-match accuracy.

\textbf{IFEval}~\citep{zhou2023instruction} evaluates the ability of language models to follow explicitly verifiable instructions. It contains approximately 500 prompts spanning 25 instruction types, with compliance automatically evaluated at both the prompt and instruction levels.

\textbf{WildGuardTest}~\citep{han2024wildguard} is a human-annotated safety moderation benchmark containing 5,299 examples with labels for prompt harmfulness, response harmfulness, and response refusal. We use its harmful prompts to evaluate the harmfulness of model-generated responses.

\textbf{HarmBench}~\citep{pmlr-v235-mazeika24a} is a standardized benchmark for evaluating automated red teaming and robust refusal against a diverse set of harmful behaviors. We use its textual behaviors directly as test prompts to measure attack success rates.

\textbf{XSTest}~\citep{rottger2024xstest} is a diagnostic benchmark for exaggerated safety behavior, comprising 250 safe prompts and 200 unsafe contrast prompts. It evaluates whether models appropriately comply with safe requests while refusing unsafe ones.

\textbf{DoAnythingNow}~\citep{shen2024anything} evaluates model robustness against in-the-wild jailbreak attacks. It analyzes 1,405 jailbreak prompts and identifies 131 jailbreak communities exhibiting diverse attack strategies.

\textbf{OCRBench}~\citep{liu2024ocrbench} is a comprehensive benchmark for evaluating the OCR capabilities of large multimodal models. It contains 1,000 manually verified question-answer pairs drawn from 29 datasets and covers text recognition, scene-text VQA, document-oriented VQA, key information extraction, and handwritten mathematical expression recognition.

\textbf{MMSI-Bench}~\citep{yang2025mmsi} is a visual question answering benchmark for multi-image spatial intelligence. It contains 1,000 expert-constructed multiple-choice questions selected from a pool of more than 120,000 images and evaluates spatial grounding and reasoning across multiple views.

\textbf{EmbSpatial-Bench}~\citep{du2024embspatial} evaluates spatial understanding in embodied environments. It is automatically derived from embodied scenes and covers six egocentric spatial relationships.

\textbf{MMMU-medical}~\citep{yue2024mmmu} denotes the medical subset of MMMU, covering expert-level medical knowledge and multimodal reasoning. It requires models to interpret specialized visual information and answer domain-specific questions.

\textbf{PathVQA}~\citep{he2020pathvqa} is a medical visual question answering benchmark containing 32,799 manually verified question-answer pairs over 4,998 pathology images collected from textbooks and digital libraries. It includes both yes/no and open-ended questions; in our experiments, we report accuracy on the yes/no subset.

\section{Merging Methods}\label{app:merge_method}

\textbf{Task Arithmetic} (TA)~\citep{ilharco2023taskarithmetic} represents each task by the parameter difference between a fine-tuned model and its base model, referred to as a task vector, and merges multiple task vectors through linear addition. It provides a simple additive baseline for model merging.

\textbf{TIES-Merging}~\citep{yadav2023ties} mitigates interference between task vectors through three steps: trimming redundant parameters, electing the sign with the greatest total magnitude, and merging only the sign-consistent updates. It is designed to reduce destructive interference among task-specific updates.

\textbf{DARE}~\citep{yu2024dare} reduces task interference by randomly dropping a portion of parameters from each task vector and rescaling the remaining updates to preserve their expected magnitude. The resulting sparse task vectors are then merged using a standard merging strategy.

\textbf{Task Singular Vector Merging} (TSVM)~\citep{gargiulo2025tsv} exploits the matrix structure of task updates by decomposing them into singular vectors. It selects and combines task-specific singular components to reduce interference while preserving important update directions.

\textbf{OrthoMerge}~\citep{yang2026orthomerge} argues that conventional Euclidean merging can destroy the intrinsic geometric structure of pretrained weights. It therefore performs merging on the Riemannian manifold of the orthogonal group by mapping orthogonal transformations to the Lie algebra, where they can be efficiently integrated while preserving their geometric structure.

\section{Evaluation}\label{app:implementation_details}

For evaluation, we follow the protocols used by OrthoMerge throughout all experiments. General text benchmarks evaluated with lm-evaluation-harness~\citep{biderman2024lessons} use greedy decoding for generation-based tasks and likelihood-based scoring for multiple-choice tasks.
HumanEval+ and MBPP+ are evaluated separately with bigcode-evaluation-harness~\citep{allal2022framework}, using a temperature of $0.2$ and generating $10$ samples per problem. For safety evaluation, we use the Ai2 safety-eval framework on WildGuardTest, HarmBench, XSTest, and DoAnythingNow, with model responses judged by WildGuard under the default evaluation protocol.
Vision-language benchmarks are evaluated with lmms-eval~\citep{zhang2025lmms} using the Qwen2.5-VL wrapper and greedy generation. Option-based tasks are scored from the answer letters generated by the model, whereas open-ended outputs are normalized and evaluated using the corresponding benchmark-specific post-processing procedures.

We use a fixed expert ordering across all layers and both component branches within each experimental setting. For the LoRA and OFT settings, the ordering is Social-IQA, Magicoder, CommonsenseQA, NuminaMath, and ScienceQA. For the fully fine-tuned language setting, the ordering is Multilingual, Coding, Math, Instruction, and Safety. For the vision-language setting, the ordering is Optical Character Recognition, Spatial Reasoning, and Medical Multimodal Reasoning. All experiments were conducted on NVIDIA RTX A6000 GPUs with $48$\,GB of memory per GPU.

\section{Geometric Derivation of the Base-Induced Orthogonal Decomposition}
\label{app:decomposition}

\subsection{Within-Frame Reorientation and the Orientation Component}
\label{app:reorientation_component}

We derive the first-order geometry of within-frame reorientation and the
closed-form projection of a task update onto the corresponding orientation
subspace.

\paragraph{First-order geometry of within-frame reorientation.}
Consider a local in-frame change of the base-induced orthonormal frame
$Q$. Its first-order form can be written as
\begin{equation}
Q' = Q + \varepsilon QA + O(\varepsilon^2),
\label{eq:app_local_frame_change}
\end{equation}
where $\varepsilon$ denotes the magnitude of the local change and
$A\in\mathbb{R}^{n\times n}$ describes how the directions within the
frame are mixed. Requiring the perturbed frame to remain orthonormal gives
\begin{align}
{Q'}^\top Q'
&=
\left(Q+\varepsilon QA\right)^\top
\left(Q+\varepsilon QA\right)
+O(\varepsilon^2)
\nonumber\\
&=
I+\varepsilon\left(A^\top+A\right)+O(\varepsilon^2).
\label{eq:app_first_order_orthogonality}
\end{align}
Hence, preserving orthonormality to first order requires
\begin{equation}
A^\top=-A.
\label{eq:app_skew_condition}
\end{equation}
Thus, first-order reorientation within the base-induced frame is
characterized by skew-symmetric coefficients.

\paragraph{Projection onto the orientation subspace.}
For an observed task update $\Delta_i$, the within-frame coordinates are
$G_i=Q^\top\Delta_i$. The update itself is not assumed to arise from an
orthogonal transformation. Instead, we identify the part of $\Delta_i$
that lies in the orientation subspace $\mathcal T_Q$ characterized above.
Since these directions have the form $QA$ with $A^\top=-A$, this component
is obtained from
\begin{equation}
A_i^\star
=
\arg\min_{A^\top=-A}
\left\|
\Delta_i-QA
\right\|_F^2.
\label{eq:app_reorientation_projection}
\end{equation}
Using
$\Delta_i=QQ^\top\Delta_i+(I-QQ^\top)\Delta_i$
and $G_i=Q^\top\Delta_i$, the objective becomes
\begin{equation}
\left\|
\Delta_i-QA
\right\|_F^2
=
\left\|
G_i-A
\right\|_F^2
+
\left\|
(I-QQ^\top)\Delta_i
\right\|_F^2.
\label{eq:app_projection_split}
\end{equation}
The second term is independent of $A$, so the projection depends only on
the within-frame coordinates $G_i$. Since
\begin{equation}
G_i
=
\operatorname{skew}(G_i)
+
\operatorname{sym}(G_i),
\end{equation}
and the skew-symmetric and symmetric parts are orthogonal under the
Frobenius inner product, the closest skew-symmetric matrix to $G_i$ is
\begin{equation}
A_i^\star
=
\operatorname{skew}(G_i).
\label{eq:app_optimal_skew}
\end{equation}
The resulting projection of $\Delta_i$ onto $\mathcal T_Q$ is therefore
\begin{equation}
T_i
=
QA_i^\star
=
Q\operatorname{skew}(G_i)
=
Q\operatorname{skew}
\left(
Q^\top\Delta_i
\right),
\label{eq:app_reorientation_component}
\end{equation}
which gives the orthogonal projection of $\Delta_i$ onto the orientation
subspace $\mathcal T_Q$.

\subsection{Characterization of the Structural Subspace}
\label{app:structure_altering_subspace}

We characterize the orthogonal complement of $\mathcal{T}_Q$ and its
within-frame and out-of-frame parts.

The orientation subspace
$\mathcal{T}_Q=\{QA:A^\top=-A\}$
is a linear subspace of $\mathbb{R}^{m\times n}$. Under the Frobenius
inner product, its orthogonal complement is
\begin{equation}
\mathcal{T}_Q^\perp
=
\{X\in\mathbb{R}^{m\times n}:Q^\top X\ \text{is symmetric}\}
=
\mathcal{R}_Q,
\label{eq:app_orthogonal_complement}
\end{equation}
because
$\langle X,QA\rangle_F=\langle Q^\top X,A\rangle_F$
vanishes for every skew-symmetric $A$ if and only if $Q^\top X$ is
symmetric.

The internal structure of $\mathcal{R}_Q$ follows by writing any
$X\in\mathcal{R}_Q$ as $X=Q(Q^\top X)+(I-QQ^\top)X$.
Since $S=Q^\top X$ is symmetric, the first term has the form $QS$,
whereas the second is orthogonal to the column space of $Q$. Conversely,
any sum of these two forms has symmetric in-frame coordinates and therefore
belongs to $\mathcal{R}_Q$. Hence,
\begin{equation}
\mathcal{R}_Q
=
\{QS:S^\top=S\}
\oplus^\perp
\{X:Q^\top X=0\},
\label{eq:reshaping_split_revised}
\end{equation}
where the two subspaces are orthogonal because
$\langle QS,X\rangle_F=\langle S,Q^\top X\rangle_F=0$
whenever $Q^\top X=0$.


\subsection{Existence, Uniqueness, and Cross-Component Orthogonality}
\label{app:unified_decomposition_proof}

We establish that every task update admits a unique decomposition into
$T_i$ and $R_i$, and that the two component families are orthogonal across
experts, as stated in Proposition~\ref{prop:base_decomposition_revised}.

\paragraph{Existence of the decomposition.}
For any task update $\Delta_i$, its coordinates in the base-induced frame are
$G_i=Q^\top\Delta_i$. Using $I=QQ^\top+(I-QQ^\top)$,
we write
\begin{equation}
\Delta_i
=
QG_i+(I-QQ^\top)\Delta_i.
\label{eq:app_unified_frame_split}
\end{equation}
Decomposing the coordinate matrix into its skew-symmetric and symmetric parts,
$G_i=\operatorname{skew}(G_i)
+\operatorname{sym}(G_i)$,
and substituting into Eq.~\eqref{eq:app_unified_frame_split} gives
\begin{align}
\Delta_i
&=
Q\operatorname{skew}(G_i)
+
\left[
Q\operatorname{sym}(G_i)
+
(I-QQ^\top)\Delta_i
\right].
\label{eq:app_unified_constructive_decomposition}
\end{align}
Accordingly, define
\begin{equation}
T_i
=
Q\operatorname{skew}(G_i),
\qquad
R_i
=
Q\operatorname{sym}(G_i)
+
(I-QQ^\top)\Delta_i.
\label{eq:app_unified_components}
\end{equation}

Because $\operatorname{skew}(G_i)$ is skew-symmetric,
$T_i\in\mathcal{T}_Q$. Moreover, since $Q^\top Q=I$ and
$Q^\top(I-QQ^\top)=0$,
\begin{equation}
Q^\top R_i
=
\operatorname{sym}(G_i),
\label{eq:app_structural_membership}
\end{equation}
which is symmetric. Hence $R_i\in\mathcal{R}_Q$. Therefore,
Eq.~\eqref{eq:app_unified_constructive_decomposition} gives a valid
decomposition $\Delta_i=T_i+R_i$ for every task update.

\paragraph{Uniqueness of the decomposition.}
Suppose that $\Delta_i$ admits two decompositions,
\begin{equation}
\Delta_i
=
T_i+R_i
=
T_i'+R_i',
\label{eq:app_unified_two_decompositions}
\end{equation}
with $T_i,T_i'\in\mathcal{T}_Q$ and
$R_i,R_i'\in\mathcal{R}_Q$. Rearranging gives
\begin{equation}
E
:=
T_i-T_i'
=
R_i'-R_i.
\label{eq:app_unified_common_difference}
\end{equation}
The left-hand side implies $E\in\mathcal{T}_Q$, while the right-hand side
implies $E\in\mathcal{R}_Q=\mathcal{T}_Q^\perp$. Thus $E$ is orthogonal to
itself, so
$\|E\|_F^2=0$,
and therefore $E=0$. It follows that $T_i=T_i'$ and $R_i=R_i'$, proving
uniqueness.

\paragraph{Cross-component orthogonality.}
Because all experts share the same reference frame $Q$, their components lie
in the same pair of orthogonal subspaces:
$T_i\in\mathcal{T}_Q$,
$R_j\in\mathcal{R}_Q=\mathcal{T}_Q^\perp$.
Hence, for any pair of experts $i,j$,
\begin{equation}
\langle T_i,R_j\rangle_F
=
0.
\label{eq:app_unified_cross_component_orthogonality}
\end{equation}
Thus, the orientation and structural component families remain orthogonal
even across different experts.
Taking $j=i$ and using $\Delta_i=T_i+R_i$ yields
\begin{equation}
\|\Delta_i\|_F^2
=
\|T_i\|_F^2
+
\|R_i\|_F^2.
\label{eq:app_unified_energy_decomposition}
\end{equation}
More generally, for any pair of experts $i,j$,
\begin{equation}
\langle \Delta_i,\Delta_j\rangle_F
=
\langle T_i,T_j\rangle_F
+
\langle R_i,R_j\rangle_F,
\label{eq:app_unified_inner_product_decomposition}
\end{equation}
because the cross terms
$\langle T_i,R_j\rangle_F$ and
$\langle R_i,T_j\rangle_F$
both vanish.

\subsection{Wide-Matrix Case}
\label{app:wide_decomposition}

For a full-row-rank weight $W_0\in\mathbb{R}^{m\times n}$ with $m<n$,
we apply the column-space construction to $W_0^\top$ using its reduced QR
factorization:
\begin{equation}
W_0^\top
=
QH,
\qquad
Q^\top Q=I,
\label{eq:app_wide_qr}
\end{equation}
where $Q\in\mathbb{R}^{n\times m}$. The columns of $Q$ form an
orthonormal frame for the row space of $W_0$.

For a task update $\Delta_i\in\mathbb{R}^{m\times n}$, its coordinates
within this row-space frame are
\begin{equation}
G_i
=
\Delta_iQ.
\label{eq:app_wide_coordinates}
\end{equation}
Applying the projections derived above to $\Delta_i^\top$ and transposing
the results back gives
\begin{equation}
T_i
=
\operatorname{skew}(G_i)Q^\top
=
\operatorname{skew}
\left(
\Delta_iQ
\right)
Q^\top.
\label{eq:app_wide_orientation_component}
\end{equation}

The structural component is correspondingly
\begin{align}
R_i
&=
\Delta_i-T_i
\nonumber\\
&=
\operatorname{sym}(G_i)Q^\top
+
\Delta_i\left(I-QQ^\top\right).
\label{eq:app_wide_structural_component}
\end{align}
Since transposition preserves the Frobenius inner product, the projection,
uniqueness, and cross-component orthogonality results established in
Appendices~\ref{app:reorientation_component}--\ref{app:unified_decomposition_proof}
carry over directly to this row-space construction.

\section{\methodname Algorithm and Implementation Details}\label{app:diga}

\subsection{Merging Procedure}\label{app:basin_algorithm}

\begin{algorithm}[H]
\caption{\methodname for a Target Matrix Parameter}
\label{alg:basin}
\begin{algorithmic}[1]

\REQUIRE Base weight $W_0$, expert weights $\{W_i\}_{i=1}^{N}$,
numerical tolerance $\epsilon$
\ENSURE Merged weight $W^\star$

\STATE $\Delta_i \leftarrow W_i-W_0$ for all $i$

\IF{$\mathrm{rows}(W_0)\geq\mathrm{cols}(W_0)$}
    \STATE $Q,H\leftarrow\mathrm{QR}(W_0)$;
    $G_i\leftarrow Q^\top\Delta_i$;
    $T_i\leftarrow Q\,\mathrm{skew}(G_i)$
\ELSE
    \STATE $Q,H\leftarrow\mathrm{QR}(W_0^\top)$;
    $G_i\leftarrow\Delta_iQ$;
    $T_i\leftarrow\mathrm{skew}(G_i)Q^\top$
\ENDIF

\STATE $R_i\leftarrow\Delta_i-T_i$ for all $i$

\FOR{$(b,\{D_i\}_{i=1}^{N})\in
\big\{(T,\{T_i\}_{i=1}^{N}),(R,\{R_i\}_{i=1}^{N})\big\}$}

    \STATE $n_i\leftarrow\|D_i\|_F$,
    $u_i\leftarrow D_i/n_i$ for all $i$;
    $\mathcal{A}\leftarrow\varnothing$

    \FOR{$i=1$ to $N$}
        \STATE $r_i\leftarrow
        u_i-\sum_{j\in\mathcal{A}}
        \langle u_i,e_j\rangle_F e_j$

        \IF{$\|r_i\|_F>\epsilon$}
            \STATE $e_i\leftarrow r_i/\|r_i\|_F$;
            $\mathcal{A}\leftarrow\mathcal{A}\cup\{i\}$
        \ENDIF
    \ENDFOR

    \STATE $v\leftarrow\sum_{i\in\mathcal{A}}n_i e_i$;
    $\hat v\leftarrow v/\|v\|_F$

    \STATE $\bar n\leftarrow\frac{1}{N}\sum_i n_i$;
    $\bar c\leftarrow
    \frac{1}{N(N-1)}
    \sum_{i\neq j}\langle u_i,u_j\rangle_F$

    \STATE $\Delta_b\leftarrow(1+\bar c)\bar n\,\hat v$

\ENDFOR

\STATE $W^\star\leftarrow W_0+\Delta_T+\Delta_R$
\STATE \textbf{return} $W^\star$

\end{algorithmic}
\end{algorithm}

\paragraph{Numerical handling.}

We use a numerical tolerance of $\epsilon=10^{-8}$ in the
Gram--Schmidt procedure. If $\|r_i\|_F\leq\epsilon$, the corresponding residual direction is treated as numerically redundant and omitted from the directional construction.
To verify that the retained directions are not near-degenerate before
normalization, Table~\ref{tab:gs_residual_norms} reports the distribution
of the pre-normalization residual norms $\rho_i=\|r_i\|_F$ for OFT and
LoRA. All measured residual norms exceed $10^{-1}$, and no direction is
removed by the tolerance. This suggests that the normalization step
operates away from the near-degenerate regime in these experiments.

\begin{table}[h]
\centering
\caption{Distribution of pre-normalization Gram--Schmidt residual norms
$\rho_i=\|r_i\|_F$ in the five-expert OFT and LoRA settings. Each row
summarizes 896 noninitial Gram--Schmidt steps across 224 target matrices
using the expert ordering of the corresponding main experiment. Values
are rounded to four decimal places.}
\label{tab:gs_residual_norms}
\small
\setlength{\tabcolsep}{3.5pt}
\begin{tabular}{llcccc}
\toprule
Adapt. & Branch & Min. & 1st pct. & 5th pct. & Median \\
\midrule
OFT  & $T$ & 0.6330 & 0.7455 & 0.8803 & 0.9971 \\
OFT  & $R$ & 0.6390 & 0.7454 & 0.8815 & 0.9971 \\
LoRA & $T$ & 0.8259 & 0.9264 & 0.9895 & 1.0000 \\
LoRA & $R$ & 0.8369 & 0.9246 & 0.9909 & 1.0000 \\
\bottomrule
\end{tabular}
\end{table}

\begin{table*}[t]
\centering
\caption{
Sensitivity of \methodname to all $5!=120$ expert orderings in the
five-expert OFT setting. OrthoMerge is included as a baseline, while
symmetric orthogonalization serves as an order-independent reference.
Mean and standard deviation are computed over all permutations.
}
\label{tab:order_sensitivity}

\resizebox{\textwidth}{!}{
\begin{tabular}{lcccccc|ccc}
\toprule
Reference / Statistic
& MATH500 & HumanEval+ & ScienceQA & CommonsenseQA & Social-IQA
& Task Avg. & AGIEval & ARC-Avg & Transfer Avg. \\
\midrule

OrthoMerge
& 24.80 & 38.41 & 87.72 & 80.51 & 55.17 & 57.32
& 38.78 & 42.75 & 40.76 \\

Symmetric orthogonalization
& 29.40 & 39.76 & 86.51 & 80.34 & 55.12 & 58.23
& 38.59 & 42.83 & 40.71 \\

\midrule

Default order
& 29.40 & 40.12 & 88.26 & 79.93 & 55.27 & 58.60
& 38.97 & 42.86 & 40.92 \\

Mean $\pm$ std
& $28.34\pm0.85$ & $39.84\pm0.53$ & $86.65\pm1.29$
& $80.29\pm0.42$ & $55.06\pm0.18$ & $58.03\pm0.31$
& $38.40\pm0.43$ & $42.88\pm0.07$ & $40.64\pm0.23$ \\

Min--max
& 25.40--30.60 & 38.60--41.28 & 84.58--88.40
& 79.52--81.24 & 54.66--55.53 & 57.04--58.67
& 37.70--39.21 & 42.68--43.01 & 40.27--41.11 \\

\bottomrule
\end{tabular}
}
\end{table*}

\subsection{Sensitivity to Expert Ordering}
\label{app:gs_order_sensitivity}

Sequential Gram--Schmidt removes from each expert direction the components
already represented by preceding experts, making the resulting basis
dependent on expert ordering. We therefore evaluate all $5!=120$
permutations of the five OFT experts.
Table~\ref{tab:order_sensitivity} reports the default ordering together
with the mean, standard deviation, and range over all permutations.
Complete per-order results are provided in
Tables~\ref{tab:perm1}, \ref{tab:perm2} and~\ref{tab:perm3}, where \textit{code},
\textit{math}, \textit{science}, \textit{comm}, and \textit{social}
denote the Magicoder, NuminaMath, ScienceQA, CommonsenseQA, and
Social-IQA experts, respectively.

The aggregate target-task performance is relatively stable across
orderings, with a Task Avg. of $58.03\pm0.31$ and a range of
$57.04$--$58.67$. Moreover, 118 of the 120 orderings achieve a higher
Task Avg. than OrthoMerge, showing that the improvement is not confined
to a small subset of favorable permutations. Order sensitivity is more
visible on individual benchmarks, particularly MATH500 and ScienceQA,
but is reduced after averaging across the five target tasks. Transfer
performance is similarly stable, with a Transfer Avg. of
$40.64\pm0.23$ and a range of $40.27$--$41.11$.

As an order-independent reference, we replace sequential Gram--Schmidt
with symmetric orthogonalization while retaining the same component-wise
aggregation procedure. This variant achieves a Task Avg. of $58.23$ and
a Transfer Avg. of $40.71$, compared with $57.32$ and $40.76$ for
OrthoMerge. These results indicate that expert ordering affects the
specific solution produced by sequential Gram--Schmidt, while the overall
target-task gains remain robust across orderings.

\begin{figure*}[h]
\centering
\includegraphics[width=\linewidth]{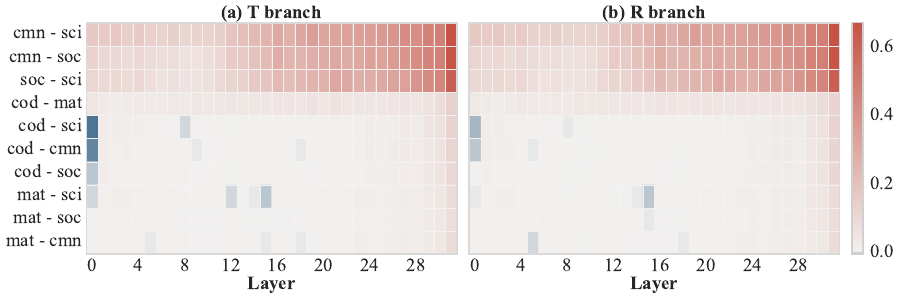}
\caption{Layer-wise pairwise cosine similarities among the orientation (a) and structural (b) components of five OFT experts. For each expert pair and layer, cosine similarity is computed separately for each attention and MLP projection matrix and averaged across projections. Abbreviations denote
commonsense (cmn), science (sci), social reasoning (soc), code generation (cod), and mathematics (mat).}
\label{fig:cos}
\end{figure*}

\begin{figure*}[t]
\centering
\includegraphics[width=\linewidth]{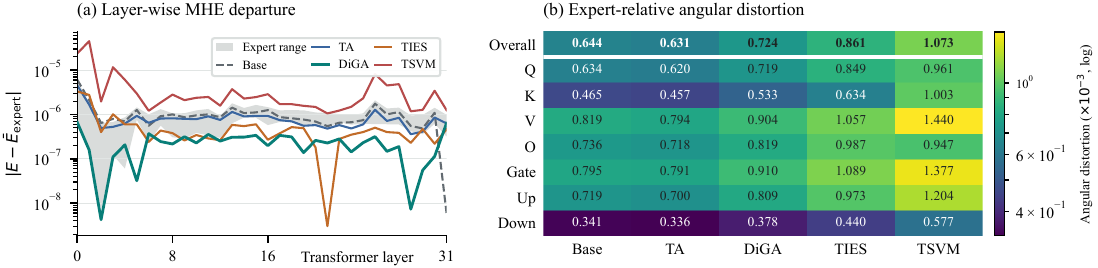}
\caption{\textbf{Geometric effects of merging on LoRA experts.}
(a) Layer-wise MHE departure from the expert regime.
(b) Expert-relative angular distortion overall and across projection types.}
\label{fig:geometric_effects_lora}
\end{figure*}

\begin{table*}[h]
\centering
\caption{Geometric statistics of the orientation and structural components for OFT and LoRA experts. $t/r$ denotes the norm ratio $\|T_i\|_F/\|R_i\|_F$. $\cos_T$ and $\cos_R$ denote the cosine similarities between the task update $\Delta_i$ and its orientation component $T_i$ and structural component $R_i$, respectively, with $\theta_T$ and $\theta_R$ denoting the corresponding angles. $t_{\mathrm{cos}}$ denotes the cosine similarity between $T_i$ and $R_i$. For each expert, all quantities are computed after flattening and concatenating the target weight matrices.}
\label{tab:oft_lora_geometry}
\resizebox{\linewidth}{!}{
\begin{tabular}{c|cccccc|cccccc}
\toprule
&
\multicolumn{6}{c|}{\textbf{OFT}}
&
\multicolumn{6}{c}{\textbf{LoRA}}
\\
\cmidrule(lr){2-7}
\cmidrule(lr){8-13}

Model
&
$t_{\cos}$&$t/r$&$\cos_T$&$\cos_R$&$\theta_T$&$\theta_R$
&$t_{\cos}$&$t/r$&$\cos_T$&$\cos_R$&$\theta_T$&$\theta_R$\\
\midrule

Commonsense
&0.000015&0.773&0.611&0.791&52.304&37.695&
0.000193&0.506&0.452&0.892&63.148&26.841\\

Magicoder
&-0.000013&0.822&0.635&0.773&50.594&39.406
&0.000007&0.531&0.469&0.883&62.024&27.975\\

NuminaMath
&-0.000017&0.866&0.655&0.756&49.112&40.889
&0.000009&0.508&0.453&0.892&63.087&26.912\\

ScienceQA
&0.000024&0.776&0.613&0.790&52.178&37.821&
0.000202&0.512&0.456&0.890&62.877&27.112\\

SocialIQA
&-0.000003&0.807&0.628&0.778&51.104&38.897
&0.000062&0.541&0.476&0.880&61.601&28.396\\

\midrule

Average
&- &0.809&0.628&0.778&51.058&38.942&
-&0.520&0.461&0.887&62.547&27.447\\

\bottomrule
\end{tabular}
}
\end{table*}


\begin{table}[h]
\centering
\caption{Performance comparison on MATH500 and HumanEval+.}
\label{tab:2merge}
\begin{tabular}{lccc}
\toprule
Model & MATH500 & HumanEval+ & Avg \\
\midrule
Llama-3.1-8B & 18.40\% & 22.44\% & 20.42\% \\
Individual Experts & 27.20\% & 38.78\% & 32.99\% \\
\midrule
TA & 27.00\% & 41.46\% & 34.23\% \\
TIES & 30.40\% & 43.72\% & 37.06\% \\
TSVM & 26.40\% & 43.35\% & 34.88\% \\
\methodname & 28.40\% & 46.89\% & 37.65\% \\
\bottomrule
\end{tabular}
\end{table}


\begin{table}[h]
\centering
\caption{Performance comparison on MATH500, HumanEval+, and ScienceQA.}
\label{tab:3merge}
\begin{tabular}{lcccc}
\toprule
Model & MATH500 & HumanEval+ & ScienceQA & Avg \\
\midrule
Llama-3.1-8B & 18.40\% & 22.44\% & 71.27\% & 37.37\% \\
Individual Experts & 27.20\% & 38.78\% & 91.28\% & 52.42\% \\
\midrule
TA & 26.60\% & 36.46\% & 87.95\% & 50.34\% \\
TIES & 27.40\% & 41.22\% & 74.69\% & 47.77\% \\
TSVM & 26.80\% & 40.43\% & 89.75\% & 52.33\% \\
\methodname & 27.20\% & 42.87\% & 90.83\% & 53.63\% \\
\bottomrule
\end{tabular}
\end{table}

\begin{table}[h]
\centering
\caption{Performance comparison on MATH500, HumanEval+, ScienceQA, and CommonsenseQA.}
\label{tab:4merge}
\begin{tabular}{lccccc}
\toprule
Model & MATH500 & HumanEval+ & ScienceQA & CommonsenseQA & Avg \\
\midrule
Llama-3.1-8B & 18.40\% & 22.44\% & 71.27\% & 70.60\% & 45.68\% \\
Individual Experts & 27.20\% & 38.78\% & 91.28\% & 82.56\% & 59.96\% \\
\midrule
TA & 25.20\% & 33.11\% & 85.79\% & 76.09\% & 55.05\% \\
TIES & 27.40\% & 41.59\% & 80.53\% & 74.94\% & 56.12\% \\
TSVM & 24.00\% & 38.90\% & 89.52\% & 79.61\% & 58.01\% \\
\methodname & 26.60\% & 40.67\% & 88.94\% & 79.20\% & 58.85\% \\
\bottomrule
\end{tabular}
\end{table}

\begin{table}[h]
\centering
\caption{Performance comparison on MATH500, HumanEval+, ScienceQA, CommonsenseQA, and Social-IQA.}
\label{tab:5merge}
\resizebox{\columnwidth}{!}{
\begin{tabular}{lcccccc}
\toprule
Model & MATH500 & HumanEval+ & ScienceQA & CommonsenseQA & Social-IQA & Avg \\
\midrule
Llama-3.1-8B 
& 18.40\% & 22.44\% & 71.27\% & 70.60\% & 48.26\% & 46.19\% \\

Individual Experts
& 27.20\% & 38.78\% & 91.28\% & 82.56\%
& 56.76\% & 59.32\% \\

\midrule

TA
& 25.20\% & 32.93\% & 83.45\% & 76.74\% & 51.89\% & 54.04\% \\

TIES
& 27.80\% & 40.00\% & 81.88\% & 77.07\% & 52.35\% & 55.82\% \\

TSVM
& 23.40\% & 35.91\% & 86.38\% & 80.18\% & 55.02\% & 56.18\% \\

\methodname
& 29.00\% & 40.37\% & 88.22\%
& 79.93\% & 55.12\% & 58.53\% \\

\bottomrule
\end{tabular}
}
\end{table}


\begin{table}[h]
\centering
\caption{Per-benchmark safety results for fully fine-tuned language expert merging. Arrows indicate whether lower or higher values are preferred. Safety is the macro average after converting the lower-is-better metrics as $100-\mathrm{score}$.}
\label{tab:safety_breakdown}
\resizebox{\linewidth}{!}{
\begin{tabular}{lccccc}
\toprule
Model & WildGuardTest $\downarrow$ & HarmBench $\downarrow$ & XSTest $\uparrow$ & DoAnythingNow $\downarrow$ & Safety \\
\midrule
Llama-3.2-3B & 72.90 & 73.13 & 45.33 & 69.00 & 32.58 \\
Individual Experts & 13.75 & 10.94 & 67.20 & 9.00 & 83.38 \\

\midrule

TA & 60.21 & 62.19 & 46.89 & 60.00 & 41.12 \\

OrthoMerge-C & 60.88 & 65.00 & 49.56 & 59.67 & 41.00 \\

OrthoMerge-G & 58.21 & 61.25 & 50.89 & 58.67 & 43.19 \\

TIES & 50.60 & 62.81 & 58.89 & 66.67 & 44.70 \\

OrthoMerge-C & 51.94 & 64.38 & 60.89 & 68.67 & 43.98 \\

OrthoMerge-G & 53.14 & 65.31 & 58.22 & 69.67 & 42.53 \\

TSVM & \underline{47.13} & 63.13 & \underline{62.22} & \underline{50.00} & \underline{50.49} \\

OrthoMerge-C & \underline{47.13} & 62.81 & 60.89 & \textbf{48.67} & \textbf{50.57} \\

OrthoMerge-G & 51.00 & \textbf{60.31} & 55.78 & 60.00 & 46.12 \\

\rowcolor{gray!15}
\textbf{\methodname} & \textbf{47.00} & \underline{60.63} & \textbf{65.11} & 64.00 & 48.37 \\

\bottomrule
\end{tabular}
}
\end{table}

\begin{table}[h]
\centering
\caption{Per-benchmark results for direct merging and component-separated processing in the five-expert OFT setting.}
\label{tab:tr_full_results1}
\resizebox{\linewidth}{!}{
\begin{tabular}{lccccccccc}
\toprule
Method & MATH500 & HumanEval+ & ScienceQA & CommonsenseQA & Social-IQA & Task Avg. & AGIEval & ARC-Avg & Transfer Avg. \\
\midrule
TIES & 27.80 & 40.00 & 81.88 & 77.07 & 52.35 & 55.82 & 37.83 & 41.99 & 39.91 \\
TIES (Comp.-Sep.) & 28.40 \textcolor{red}{(+0.60)} & 39.57 \textcolor{green}{(-0.43)} & 81.70 \textcolor{green}{(-0.18)} & 77.72 \textcolor{red}{(+0.65)} & 53.02 \textcolor{red}{(+0.67)} & 56.08 \textcolor{red}{(+0.26)} & 38.10 \textcolor{red}{(+0.27)} & 41.95 \textcolor{green}{(-0.04)} & 40.03 \textcolor{red}{(+0.12)} \\
\midrule
DARE & 18.80 & 31.28 & 83.72 & 80.92 & 58.03 & 54.55 & 34.70 & 38.37 & 36.53 \\
DARE (Comp.-Sep.) & 19.60 \textcolor{red}{(+0.80)} & 32.93 \textcolor{red}{(+1.65)} & 83.72 \textcolor{black}{(0.00)} & 81.08 \textcolor{red}{(+0.16)} & 58.24 \textcolor{red}{(+0.21)} & 55.11 \textcolor{red}{(+0.56)} & 34.74 \textcolor{red}{(+0.04)} & 38.58 \textcolor{red}{(+0.21)} & 36.66 \textcolor{red}{(+0.13)} \\
\midrule
TSVM & 23.40 & 35.91 & 86.38 & 80.18 & 55.02 & 56.18 & 36.82 & 42.34 & 39.58 \\
TSVM (Comp.-Sep.) & 24.80 \textcolor{red}{(+1.40)} & 36.59 \textcolor{red}{(+0.68)} & 86.02 \textcolor{green}{(-0.36)} & 80.02 \textcolor{green}{(-0.16)} & 55.32 \textcolor{red}{(+0.30)} & 56.55 \textcolor{red}{(+0.37)} & 37.64 \textcolor{red}{(+0.82)} & 42.40 \textcolor{red}{(+0.06)} & 40.02 \textcolor{red}{(+0.44)} \\
TSVM (Comp.-Joint) & 21.80 & 32.74 & 84.22 & 78.21 & 52.66 & 53.93 & 37.93 & 42.41 & 40.17 \\
\bottomrule
\end{tabular}
}
\end{table}


\begin{table*}[h]
\centering
\caption{Complete results for all 120 expert orderings in the five-expert OFT setting (Part 1 of 3).}
\includegraphics[width=\linewidth]{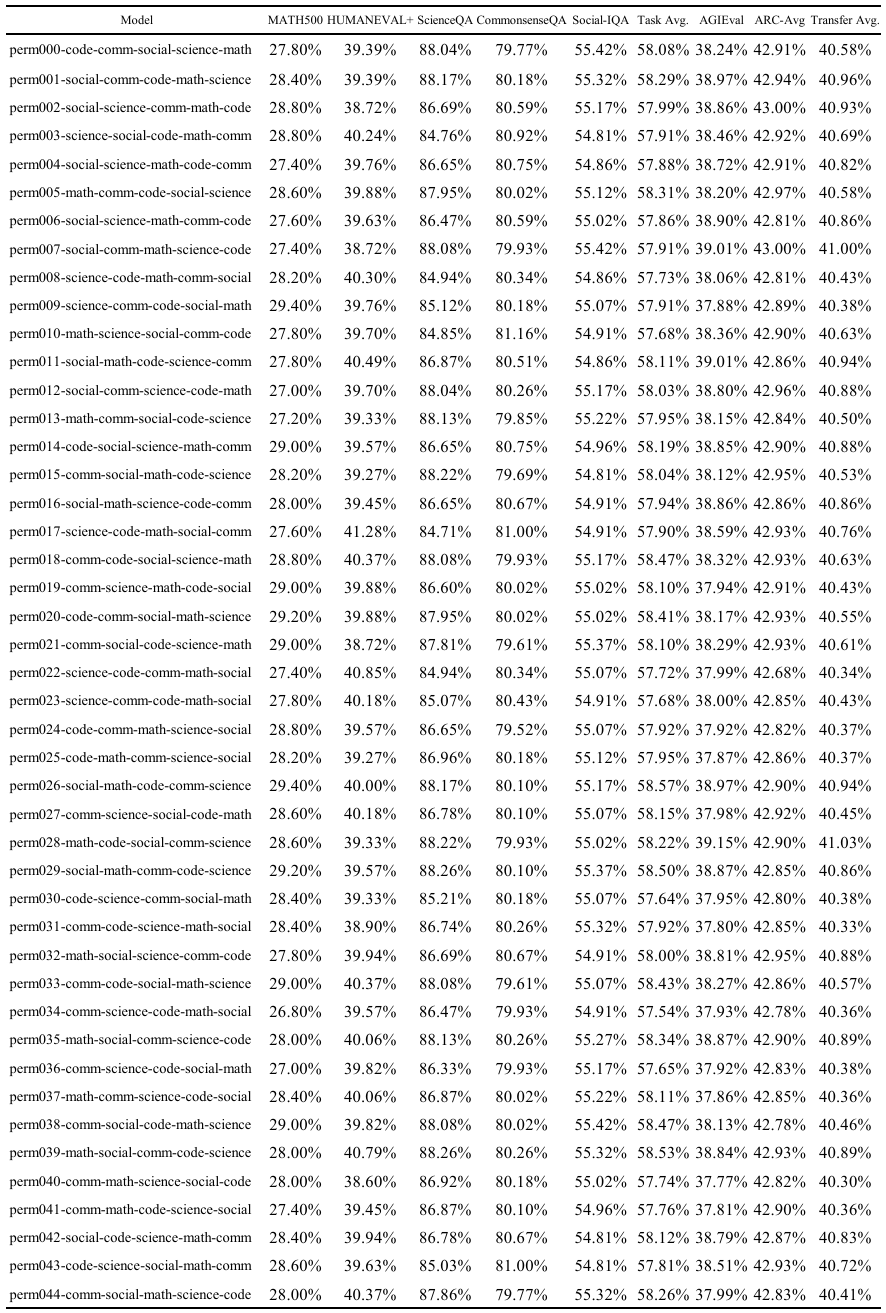}
\label{tab:perm1}
\end{table*}

\begin{table*}[t]
\centering
\caption{Complete results for all 120 expert orderings in the five-expert OFT setting (Part 2 of 3).}
\includegraphics[width=\linewidth]{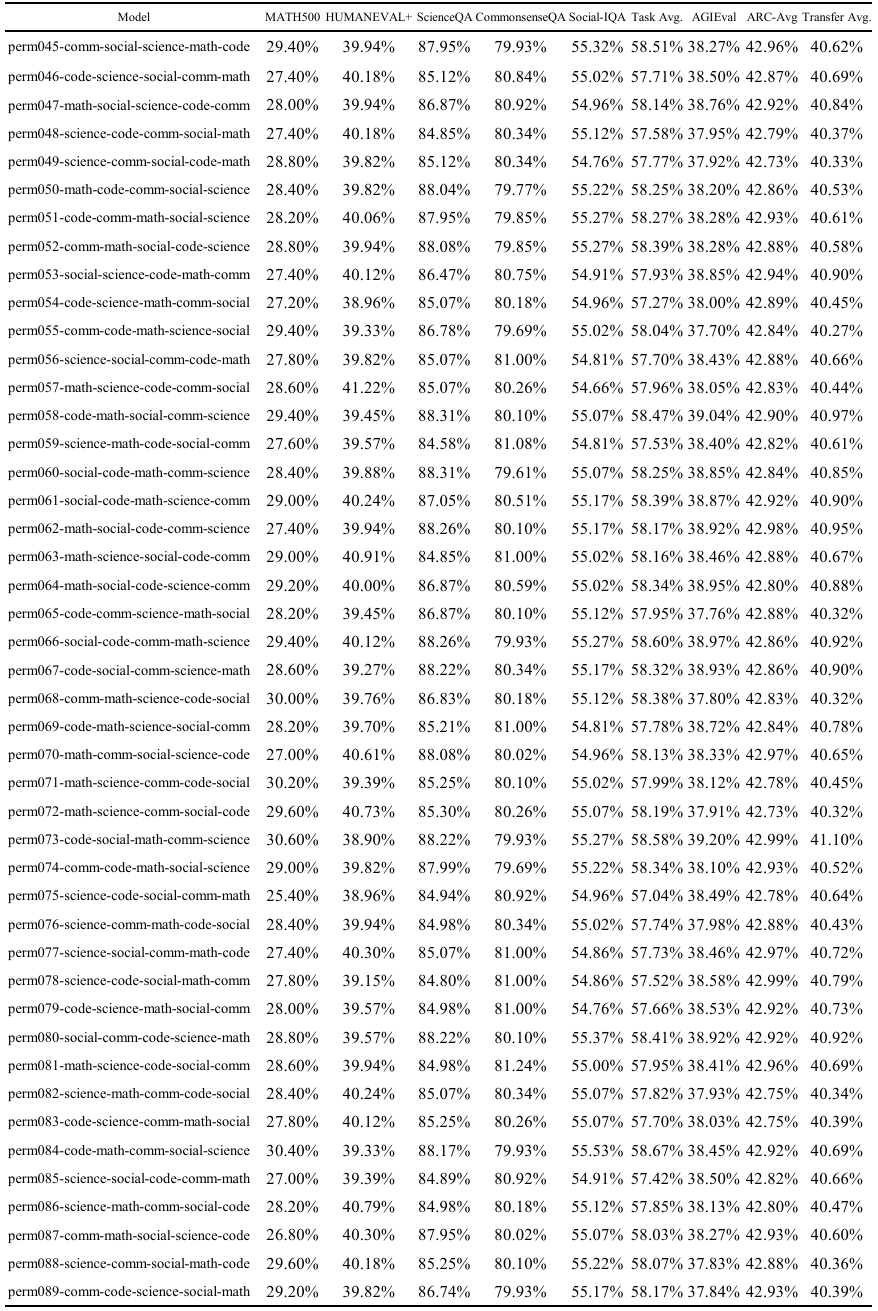}
\label{tab:perm2}
\end{table*}

\begin{table*}[t]
\centering
\caption{Complete results for all 120 expert orderings in the five-expert OFT setting (Part 3 of 3).}
\includegraphics[width=\linewidth]{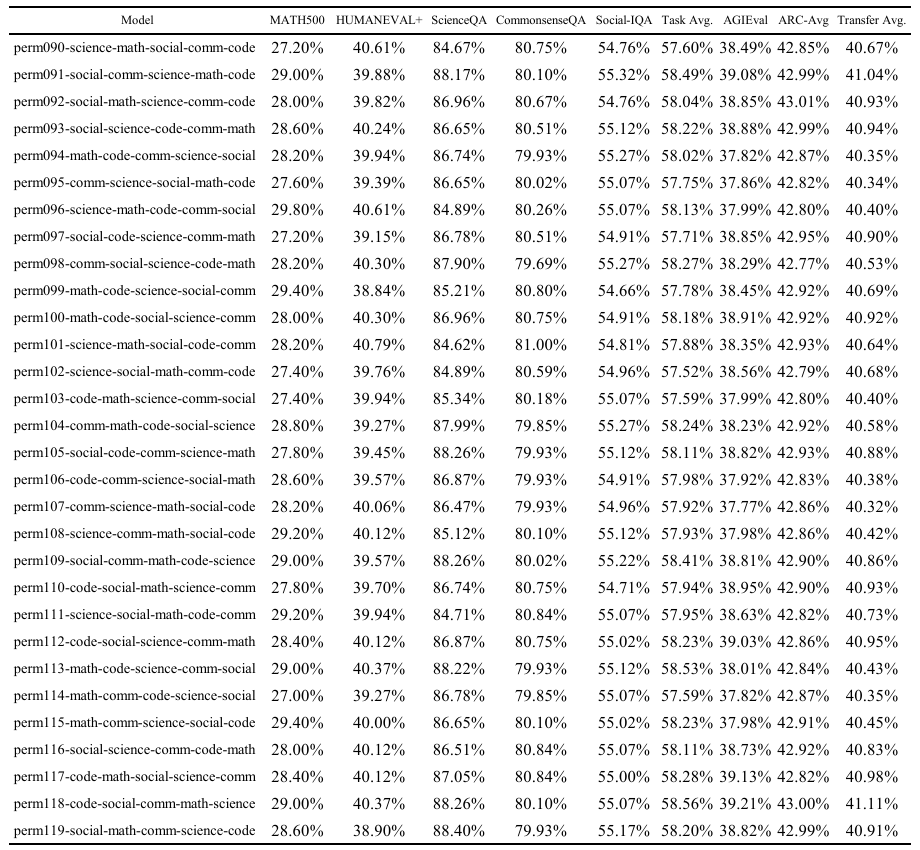}
\label{tab:perm3}
\end{table*}

\end{document}